\documentclass{ieeeaccess}
\usepackage{cite}
\usepackage{amsmath,amssymb,amsfonts}
\usepackage{algorithmic}
\usepackage{booktabs}
\usepackage{graphicx}
\usepackage{float}

\usepackage{textcomp}
\def\BibTeX{{\rm B\kern-.05em{\sc i\kern-.025em b}\kern-.08em
    T\kern-.1667em\lower.7ex\hbox{E}\kern-.125emX}}
\begin{document}
\history{}
\doi{}

\title{Palm Vein Recognition via Multi-task Loss Function and Attention Layer}

\author{\uppercase{Jiashu Lou}\authorrefmark{1}\IEEEmembership{Student member, IEEE},
\uppercase{Jie Zou\authorrefmark{2}, and Baohua Wang}.\authorrefmark{3}}
\address[1, 2, 3]{College of Mathematics and Statistics, Shenzhen University, No. 3688, Nanhai Avenue, Shenzhen, Guangdong, China}
\address[1]{loujiashu@163.com}
\address[2]{ianzou2000@163.com}
\address[3]{bhwang@szu.edu.cn}

\begin{abstract}
With the improvement of arithmetic power and algorithm accuracy of personal devices, biological features are increasingly widely used in personal identification, and palm vein recognition has rich extractable features and has been widely studied in recent years.
However, traditional recognition methods are poorly robust and susceptible to environmental influences such as reflections and noise.
In this paper, a convolutional neural network based on VGG-16
transfer learning fused attention mechanism is used as the feature extraction network on the infrared palm vein dataset.
The palm vein classification task is first trained using palmprint classification methods, followed by matching using a similarity function, in which we propose the multi-task loss function to improve the accuracy of the matching task.
In order to verify the robustness of the model, some experiments were carried out on datasets from different sources.
Then, we used K-means clustering to determine the adaptive matching threshold and finally achieved an accuracy rate of 98.89\% on prediction set. 
At the same time, the matching is with high efficiency which takes an average of 0.13 seconds per palm vein pair, and that means our method can be adopted in practice.
\end{abstract}


\titlepgskip=-15pt

\maketitle

\section{Introduction}
\label{sec:introduction}
In today's online information society, we increasingly need to prove our identity in different public places. Previously, we used traditional methods of identification such as passwords, ID cards, family registers, and driving licenses, but these methods are easy to forget, lose and steal, and they have a number of security risks. In the era of Big Data, we have high-performance computers that have the computing power to take and process unique biological characteristics as a "password" for identification, such as the face, fingerprint, palm print, iris, voice, and other biometric characteristics.\par

Among these, palmprint and palm vein recognition has become very effective technology in the field of biometric identification. In contrast to other biometric features, there are many available lines in the palm area, such as main lines, mastoid lines, and wrinkles. At the same time, palm veins have the advantages of high differentiation, stability, and low invasiveness. Currently, palm print and palm vein recognition has received a great deal of attention from researchers and is widely used in criminal investigation and public security.\par

In this paper, we propose a convolutional neural network based on VGG-16
transfer learning fused attention mechanism which is used as the feature extraction network on the infrared palm vein dataset. In the training phase, we consider the recognition of palm veins as a classification task, and in the prediction we need to match the palm veins two by two. Therefore, we introduce the multi-task loss function, which is found to be effective in improving the matching accuracy after experiments. Then, we use clustering to determine the adaptive matching threshold and finally achieve an accuracy rate of 98.89\% on prediction set.

\section{Related work}
\subsection{Palm Vein Recognition}
Biometric identification technology uses an individual's unique biometric characteristics to uniquely determine an individual's identity \cite{ref1}\cite{ref2}. With the development of mobile Internet and the increase of computing speed of personal electronic devices, technologies such as facial recognition \cite{ref5}\cite{ref6}, fingerprint recognition \cite{ref3}\cite{ref4}, and even iris recognition \cite{ref7} have been integrated into every personal terminal. Moreover, palmprint recognition as an emerging biometric technology has recently appeared in the public eye. Palmprints have richer features than fingerprints, which are not easy to change and have better identification and uniqueness. \par

Palmprint research uses high-resolution or low-resolution images. High-resolution images are suitable for forensic applications such as criminal investigation, e.g., Jain, AK (2009)\cite{ref11} developed latent palmprint matching, mainly dealing with low-resolution palmprints. Since about 30\% of the biopsies recovered from crime scenes come from the palm, the evidentiary value of palm prints in forensic applications is evident. On the other hand, low-resolution images are more suitable for civilian and commercial applications such as access control. Generally, high resolution is defined as 400 dpi or higher, and low-resolution is defined as 150 dpi or lower. Ridges, singularities, and detail points are generally extracted as features in high-resolution images. While in low-resolution images, main lines, wrinkles, and textures are usually extracted. Initially, palmprint research focused on high-resolution images \cite{ref29}\cite{ref30}. However, recent research hotspots have shifted to commercial and residential low-resolution palmprint recognition and matching \cite{ref31} \par

Palm vein features are similar to it, and its application is better \cite{ref38} because the features of veins are more evident and unique. Moreover, compared to palm prints, palm veins rely on biometric information inside the body and are therefore less susceptible to destruction, alteration, or tampering. Therefore, vein recognition is becoming one of the most reliable methods in biometrics and has attracted broad interest from biometric researchers. Veins are vast networks of blood vessels under the human skin that are almost invisible to the human eye and are more difficult to replicate than other biometric features \cite{ref40}. The shape of the vascular pattern is thought to be unique across individuals (Wilson, 2010)\cite{ref39} and stable over time.

For now, the academic community mainly focuses on the extraction and localization of palmprint features using mathematical means, such as Huang (2008) et al. \cite{ref9} and Wu (2006) et al. \cite{ref10} using intrinsic features of palmprints, such as main lines and wrinkles, for palmprint recognition.
; Lu, GM (2003) \cite{ref8} proposed a method to extract palmprint feature vectors by Karhunen-Loeve transformation; Zhang, D (2010) extended the feature space to three dimensions, extracted depth features using structured light techniques, and developed a multilevel framework for personal identity verification. Palmprint matching has been richly used in various fields. In addition, deep learning networks are gradually used in palmprint recognition with the popularity of deep learning and the increase of computing hardware arithmetic power. For example, Zhao, SP (2022) \cite{ref12} used deep convolutional neural networks to extract palmprint features and developed a joint constrained least squares regression framework for palmprint recognition; Trabelsi, S (2022) \cite{ref13} used a simplified PalmNet-Gabor method to improve PalmNet and obtained a higher high accuracy, reducing the number of features and saving computation time.

\subsection{Convolutional Neural Networks}

Convolutional neural networks (CNNs) use the original image as input, which can effectively learn the corresponding features from many samples and avoid the complicated feature extraction process. Unlike fully connected neural networks, CNNs can directly process two-dimensional images, while the former loses dimensional information in image spreading. Therefore, CNNs have been widely used in image processing. For example, R Vinoth (2014)\cite{ref27} et al. used CNN networks for tumor image recognition and segmentation in magnetic resonance imaging and compared them with traditional methods such as SVM. Xiao Huang (2019)\cite{ref28} et al. used two CNNs to extract visual and textual features from social media posts and identify disasters associated with related social media for fast response. It can be seen that convolutional neural networks have made outstanding contributions in the field of image recognition. \par

However, a deep convolutional neural network contains a large number of parameters to be trained (e.g., ResNet-50\cite{ref17} has 23 million parameters), and we need a large amount of labeled data to train it, but manually labeling the data would be a tremendous amount of work. One idea is to use unsupervised learning to annotate the data, as in Kiselev, VY (2019)\cite{ref15} Unsupervised learning of single-cell RNA-seq data using clustering. Another direction is transfer learning \cite{ref16}, which can take knowledge from one domain (source domain) and migrate it to another domain (target domain), enabling the target domain to achieve better learning results. This is especially true for scenarios where the source domain has sufficient data, and the target domain has a small amount of data. In image recognition and classification, transfer learning has achieved good performance. Shin, HC (2016)\cite{ref14} used a network pre-trained on ImageNet dataset for medical image recognition by transfer learning for thoracoabdominal lymph node (LN) detection and interstitial lung disease (ILD) classification and reported achieving the best performance. The combination of CNNs and transfer learning allows for a much broader range of deep learning applications.

\section{Methods}
The model in this paper will be divided into two aspects: training and testing,  since ultimately the task to be solved in this paper is not to classify the given palm vein images, but to determine whether the given pair of palm vein images belong to the same person. If reconstructing the dataset into pairs of palm vein images and labels of whether they belong to the same person, will bring a large workload and is not conducive to transfer learning using existing convolutional networks. Therefore, after extracting the optical features, we use VGG-16 for feature extraction, send the data to the linear classification layer during training, and use the output given by this layer to calculate the loss in combination with the training labels for training. After the training, the test data is sent into the VGG-16 feature extraction layer to extract features, and then the extracted features are spread into a feature vector, and the similarity of the two image feature vectors is calculated using the similarity function. And a certain threshold value is used to decide whether these two images belong to one person or not. Figure\ref{fig1} gives the general structure of the model in this paper, and each link will be introduced in this paper.

\begin{figure}[htbp]
    \centering
\begin{center}
    \includegraphics[width = \linewidth]{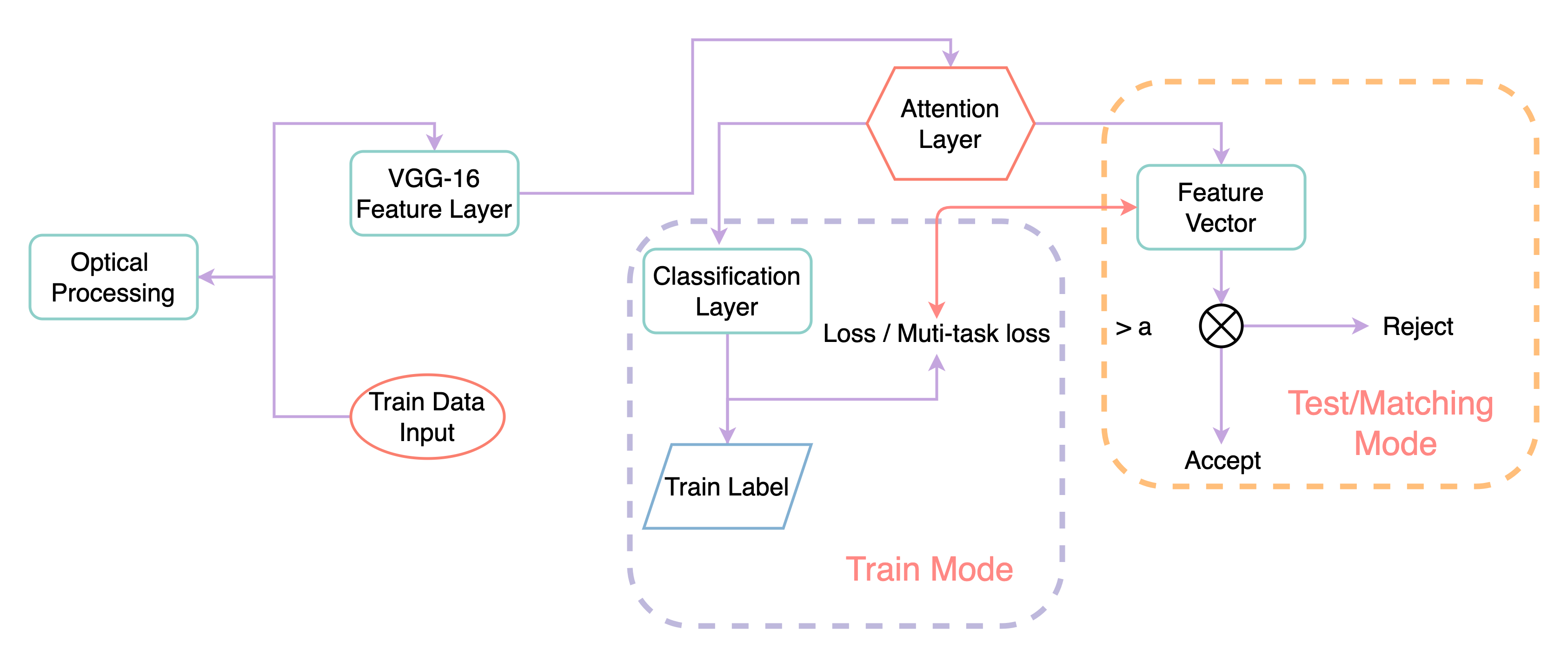}
\end{center}
    \caption{Model Structure}
    \label{fig1}
\end{figure}

\subsection{VGG Transfer Learning}
\subsubsection{Introduction of VGG structure}
K Simonyan proposed the VGG deep learning network, A Zisserman in 2014 \cite{ref18}, using an architecture with very small (3x3) convolutional filters for a comprehensive evaluation of networks with increasing depth, and introducing a 1*1 convolutional kernel in the convolutional structure of VGG, without affecting the input-output dimensionality, and introducing nonlinear transformations to increase the expressiveness of the network and reduce the computational effort. They demonstrate that increasing the depth to 16-19 weight layers can significantly improve the existing technical configuration. \par
VGG-16 has 16 layers, divided into 13 convolutional layers and 3 fully connected layers. The first time, after two convolutions of 64 convolutional kernels, one pooling is used, the second time, after two convolutions of 128 convolutional kernels, another pooling is used, and two more repetitions of three 512 convolutional kernels are convolved and then pooled. Finally, three full connected layers are used to obtain the results. The schematic structure is shown below\cite{ref51}.

\begin{figure}[htbp]
    \centering
\begin{center}
    \includegraphics[width = \linewidth]{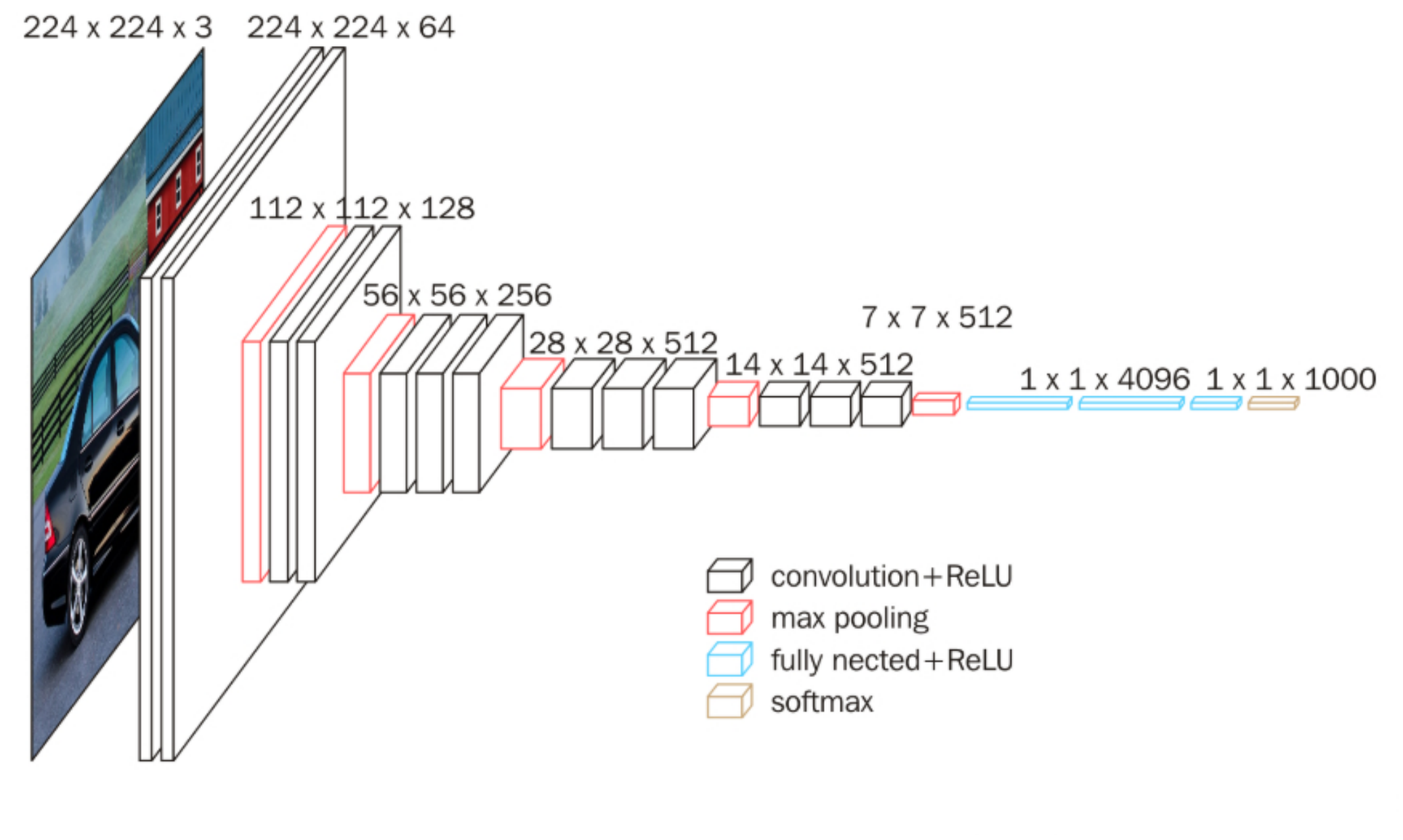}
\end{center}
    \caption{VGG-16}
\end{figure}\par

At the end of the network, VGG uses a set of fully connected nodes for the output, where the final output size is the number of categories of images in the dataset. In the middle of each fully connected layer, we add a dropout layer to randomly deactivate a certain percentage of neurons, which avoids the occurrence of model overfitting \cite{ref42}. Finally, the obtained results are fed into the $Softmax$ function (\ref{equ3}) to find the probability distribution, and finally, the loss is calculated and the error is back-propagated by the cross-entropy loss function (\ref{equ4}).
\begin{equation}
    \operatorname{Softmax}\left(z_{i}\right)=\frac{e^{z_{i}}}{\sum_{c=1}^{C} e^{z_{c}}}
    \label{equ3}
\end{equation}

\begin{equation}
    L=\frac{1}{N} \sum_{i} L_{i}=-\frac{1}{N} \sum_{i} \sum_{c=1}^{M} y_{i c} \log \left(p_{i c}\right)
    \label{equ4}
\end{equation}
The softmax function can effectively map the output to a probability distribution space and is particularly effective for solving classification problems.

\subsubsection{Transfer Learning}
As more and more machine learning application scenarios emerge and existing supervised learning that performs better requires large amounts of labeled data, so transfer learning is receiving more and more attention. The goal of transfer learning is to apply the knowledge or patterns learned on a domain or task to a different but related domain or problem \cite{ref22}. \par

For this paper's palm vein recognition task, we use a VGG-16 network pre-trained on the ImageNet\cite{ref21} dataset for transfer learning. imageNet is pre-trained on more than one million images containing more than 20,000 categories. The data volume of the palm vein dataset involved in this paper is small, because there is still some difference between object recognition and palm vein recognition. Considering the above, we freeze the first 30 convolutional layers in VGG-16 with gradient, i.e., they do not generate gradient information during the training process, thus the parameters will not be updated. Instead, the remaining 34 convolutional layers are trained with fully connected layers, which greatly shortens the training time and increases the convergence rate while ensuring prediction accuracy.
\subsubsection{Attention mechanism}
When a human sees a set of information, human neurons will automatically scan the global image based on experience and the current target, and obtain the target area that needs to be focused on, that is, the focus of attention. More attention is then devoted to this region to obtain more details about the target and to suppress other useless information. The following figure is a schematic diagram of the application of the attention mechanism in image processing. For such a journal screenshot, attention is focused on the image and the caption, respectively, under our life experience. \cite{ref23}.
\begin{figure}[htbp]
    \centering
\begin{center}
    \includegraphics[width = 0.6\linewidth]{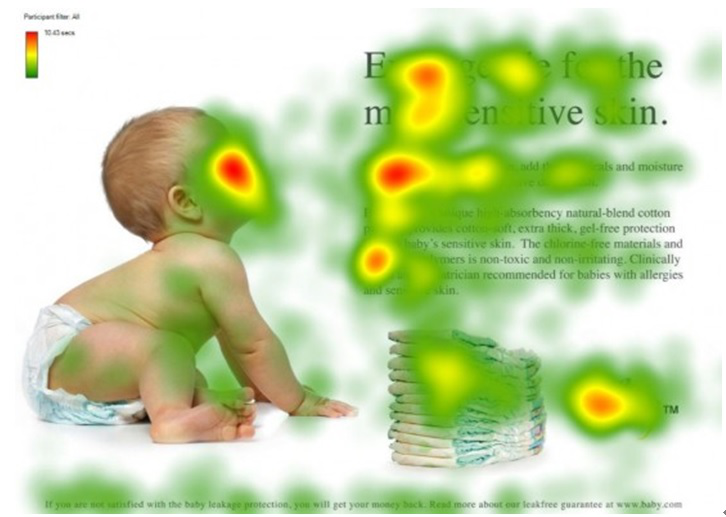}
\end{center}
    \caption{Attention}
\end{figure}\par

A brief description of the attention layer is as follows. First, two hidden states (Encoder and Decoder) are created, which are equivalent to the input and output of the structure. Subsequently, a dot product is performed with each hidden state in Decoder and Encoder, and the result is noted as $Score$. Then all the scores are sent to the softmax layer (\ref{equ3}), so that the higher the original score of the hidden state, the higher its corresponding probability, thus suppressing that invalid or noisy information. \par
Next, the hidden state of each Encoder is multiplied by the $Score$ after softmax, and then all aligned vectors are accumulated and the context vectors are sent to the Decoder to obtain the decoded output. the internal structure of the Attention layer is as follows.

\begin{figure}[htbp]
    \centering
\begin{center}
    \includegraphics[width = 0.8\linewidth]{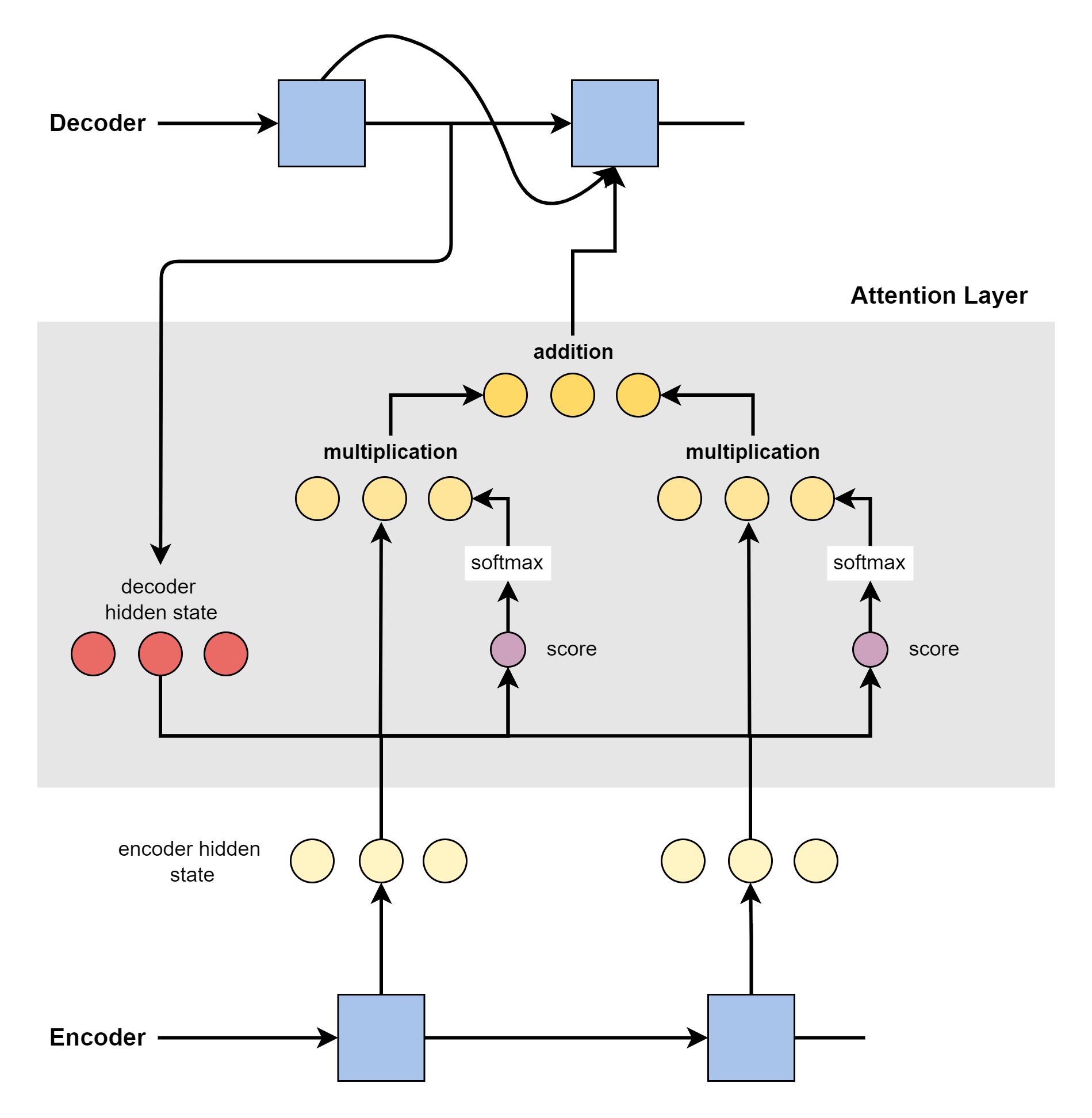}
\end{center}
    \caption{Attention layer structure}
\end{figure}\par

For the palm vein recognition task, we want the neural network to focus more on the features and patterns of the hand, and give a relatively low weight to secondary factors such as shadows and illumination. Therefore, we use the spatial Attention mechanism before image input and the channel Attention mechanism for image feature extraction \cite{ref24}. Another advantage of the Attention layer is that it is a plug-and-play adaptive module, which does not change the dimensionality of the image and does not change the subject model. It can also be applied to the training of transfer models before the input or after the output of the subject model.

\subsection{Training and matching algorithms}
\subsubsection{Framework and optimization}
In this paper, we build and train the model based on the Pytorch framework \cite{ref26}, and optimize the model using the Adam algorithm \cite{ref25}. Adam algorithm is different from the traditional stochastic gradient descent which maintains a single learning rate to update all the weights, and the learning rate does not change during the training process. In contrast, the Adam algorithm designs independent adaptive learning rates for different parameters by computing the gradient's first-order moment estimates and second-order moment estimates. This allows the loss to drop faster in the early learning period without jaggedness in the late learning period due to excessive learning rate. \par

Meanwhile, we use the (\ref{equ9})$L_2$ penalty term for the linear output layer, which enables the network to output a relatively sparse feature vector, which is more helpful for the subsequent similarity calculation. The $L_2$ penalty term can limit the size of the secondary elements in the weights, and in terms of optimization difficulty $L_2$ penalty is more convex compared to $L_0$ and $L_1$ penalties \cite{ref32}, so the optimization is less difficult and the results obtained are better. Also, the inclusion of the regular term helps to avoid overfitting because it compresses part of the parameters to a number close to 0, making the model simpler.

\begin{equation}
    penalty = \left \| w \right \|_2
    \label{equ13}
\end{equation}\par

Where $\left \| w \right \|_2$ is the L2 penalty term, which is easier to compute than the L1 parametrization, and the derivative values are defined at each point. Including the L2 parametrization is achieved simultaneously to facilitate optimisation and obtain a sparse solution. The expression of the loss function after adding the canonical term is as follows.

\begin{equation}
    loss = Crossentropy(X, y) + \lambda\left \| w \right \|_2
    \label{equ9}
\end{equation}

\subsubsection{Matching algorithm}
After feature extraction by the above deep learning network, for the input $i$th palm vein image, we can get the feature vector $V_i$. Subsequently, to determine whether the $i$th and $j$th images are from the same person, we use the cosine similarity function for matching comparison, whose expression is shown in (\ref{equ5}).

\begin{equation}
    s=\cos (\theta)=\frac{V_i \cdot V_j}{\|V_i\|\|V_j\|}=\frac{\sum_{k=1}^{n} V_{ik} \times V_{jk}}{\sqrt{\sum_{k=1}^{n}\left(V_{ik}\right)^{2}} \times \sqrt{\sum_{k=1}^{n}\left(V_{jk}\right)^{2}}}
    \label{equ5}
\end{equation}\par

This function accepts two feature vectors extracted from the neural network and obtains the similarity by employing a generalised cosine operation. When $s>\alpha$, we decide that the two palm vein pictures originate from the same person, and reject them in the opposite case. Obviously, in the most ideal case, for any $V_i$ and $V_j$ coming from the same person, there should be:
\begin{equation}
    s=\cos (V_i,V_j) = 1
    \label{equ10}
\end{equation}
And for any $V_i$ and $V_j$ from different persons, there should be:
\begin{equation}
    s=\cos (V_i,V_j) = 0
    \label{equ11}
\end{equation}\par
But in practical scenarios, we can relax this judgment condition. We usually use the threshold $\alpha$ to determine whether it is the same person. It follows that as long as the difference between the similarity $s_i$ for any pair of palm veins from the same person and the similarity $s_j$ for any pair of palm veins from different people is large enough, then we can always find a suitable $\alpha$ that allows us to distinguish whether a given input is the same person or not. That is, the difference between the similarity values computed by the similarity function from the same person and different persons should be as large as possible, i.e., equation (\ref{equ8}) should be as large as possible. We call this equation \textbf{matching goodness of fit}, which is counted as $MG$.

\begin{equation}
    MG =  Average(\cos (V_{is},V_{js})) - Average(\cos (V_{id},V_{jd}))
    \label{equ8}
\end{equation}\par

This facilitates us to pick a better threshold $\alpha$ more easily, making the rate of wrong rejections and wrong acceptance decrease.

\subsubsection{Multi-task Loss Function}

Our loss function is shown in (\ref{equ9}) for the traditional classification task. That is, we solve the problem posed by (\ref{equ8}) by solving the (\ref{equ9})-style. However, there is no research to prove that there is a relaxation optimization relationship between the two, and it may not be effective if the classification model is more indirect for the matching task. Therefore, we would like to add (\ref{equ8}) to the loss function for optimization. The reconstructed loss function is shown as follows. It can be seen as a linear combination of a classification task and a matching task.

\begin{equation}
   loss = \theta * Crossentropy(X, y)  +(1-\theta)(1-MG) + \lambda\left \| w \right \|_2
    \label{equ14}
\end{equation}\par

Although we want the value of equation (\ref{equ8}) to be the largest for all data (i.e., $1-MG$ is the smallest), it is very detrimental to training because it takes too much time to compute $MG$. However, since we divide the training batches randomly, each training batch $1-MG$ minimum is equivalent to the overall $1-MG$ minimum. \par
The parameter $\theta$ controls the weight between two tasks, $\theta = 1$ for a pure classification task and $\theta = 0$ for a pure matching task.
It is worth noting that the premise of computing the batch $MG$ is that there must be both data of the same category and data of different categories in each batch. To achieve this, we may need to increase the data in each batch, which inevitably leads to a decrease in training accuracy. We will elaborate on the specific parameter selection in Chapter 3.

\section{Experiment and Result}
The following is the training environment for this article.
\begin{itemize}
\item[$\bullet$] Python 3.9.12
\item[$\bullet$] Pytorch 1.11.0
\item[$\bullet$] CUDA 11.6
\item[$\bullet$] Intel(R) Xeon(R) Gold 5320 CPU @ 2.20GHz
\item[$\bullet$] Tesla V100*1
\item[$\bullet$] 32GB RAM
\end{itemize}

To simulate a realistic application, usually without GPU acceleration, here is the test environment for this article.

\begin{itemize}
\item[$\bullet$] Macbook pro
\item[$\bullet$] Apple silicon M1 Pro
\item[$\bullet$] 16GB RAM
\end{itemize}

\subsection{Datasets}
There are currently five commonly used 2D palmprint databases in the academic community, including the Hong Kong Polytechnic University's Palmprint II database (PolyU II), the Hong Kong Polytechnic University's Blue Ribbon Multispectral database (PolyU M\_B)\cite{ref46}, the Hefei University of Technology's Palmprint database (HFUT)\cite{ref47}, the Hefei University of Technology's Cross-Sensor Palmprint database (HFUT CS)\cite{ref48}. The Tongji University Palmprint database (TJU-P)\cite{ref49}. The dataset used in this paper is PolyU M\_B and TJU-P \cite{ref34}\cite{ref43}. The datasets from TJU-P have been centered so that all the palm vein areas are recognizable on the screen, while the PolyU M\_B datasets are the entire palm, which may make prediction more difficult, however, at the same time, more common in real life.

\begin{figure}[htbp]
    \centering
\begin{center}
    \includegraphics[width = \linewidth]{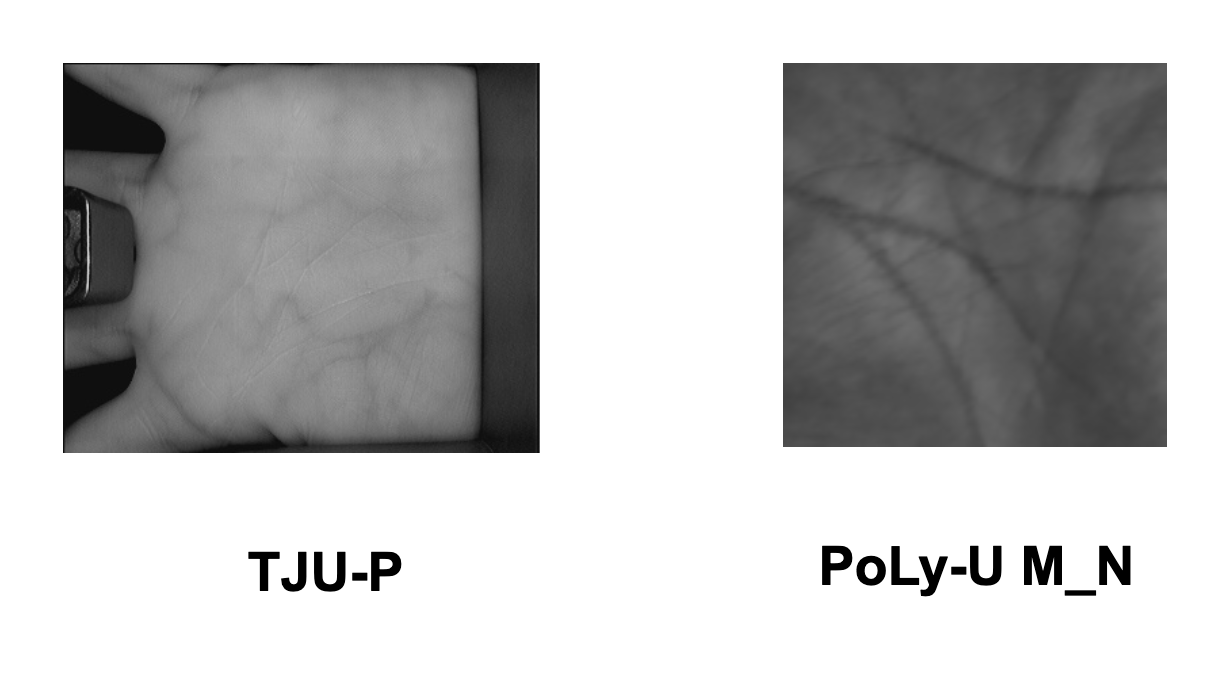}
\end{center}
    \caption{A sample from the two datasets}
    \label{fig1}
\end{figure}

These databases generally contain two parts, derived from the same samples, a period of time before and after palm print or palm vein sampling. When using deep learning methods for palm print and palm vein recognition, it is easy to improve recognition accuracy if the training set contains samples from both parts, and this experimental mode is referred to as the mixed data mode. If the training set is only from the first part and the test samples are from the second part, the training pattern is referred to as the separated data pattern\cite{ref50}. In this experiment, the separated data model was used for this task in order to obtain a more accurate and realistic recognition rate.

\subsection{Data pre-processing}
\subsubsection{Image size processing}
To fit the input size of the convolutional network, we first cropped the image to 224*224 centered on the palm to be the ROI(region of interest). The images were sourced from the same person by recording 6 consecutive images and 6 images 10 days later. That is, there are 12 pictures from the same hand each. Therefore, we use the former as the training set, counting as the set $TR$ for the classification task. And every 9 photos are taken out as the validation set, which is counted as the set $V$. For the matching process, in order to simulate the entry-comparison process in real application scenarios, we extract the last 10 sets of photos (i.e., 10 different palms) from the dataset as the test set for zero-experience learning matching to test the matching effect, and match the photos in the two datasets one by one to form a 60*60 dataset, which is counted as The set $TE_m$.

\subsubsection{Optical pre-processing}
Image enhancement is mainly used to highlight certain information or to meet specific needs. Techniques for image enhancement generally fall into two main categories: null domain image enhancement and frequency domain image enhancement algorithms, in addition to image enhancement algorithms that combine fuzzy theory and genetics.\par Space domain image enhancement is the direct processing of image pixels and is essentially based on a grey-scale mapping transformation. Commonly used methods include histogram equalization, spatial filtering, and grey scale transformations.\par The essence of frequency domain image enhancement is convolution. It mainly uses the Fourier transform or wavelet transform to transform the image to the frequency domain, then frequency domain enhancement, and finally inverse transformation back to the null domain to obtain the desired image. Common methods include low-pass filtering, high-pass filtering, etc.\par Enhancement of palm print images can eliminate noise in the original image, balance illumination effects and enhance contrast. This paper selects four image enhancement methods: histogram equalization, Laplace operator, Constrained Contrast Adaptive Histogram Equalisation algorithm (CLAHE), and log transform. Two samples were randomly selected on two datasets respectively and processed as shown in the following figures .\par

\begin{figure}[H]
	\centering
	\includegraphics[width=\linewidth]{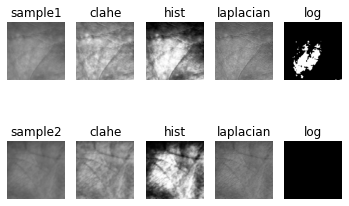}
	\caption{Impact of four optical pretreatments on the TJU-P dataset}
	\label{zq1}
\end{figure}

\begin{figure}[H]
	\centering
	\includegraphics[width=\linewidth]{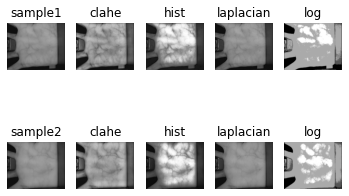}
	\caption{Impact of four optical pretreatments on the PoLy-U M\_N dataset}
	\label{zq2}
\end{figure}

Among them, the log transform was the least effective on the TJU-P dataset, while the laplacian, clahe, and hist effectively removed the noise from the original images. On the PolyU M\_N dataset, the laplacian operator processed results do not differ much from the original image, while the remaining methods each highlight distinct features after enhancement.

\subsection{Judging criteria}

For a single pair of samples, we can use the deviation from the ideal state equation (\ref{equ10}, \ref{equ11}) to measure the deviation. Since the matching result is essentially a dichotomous variable of "acceptance" and "rejection" for a given multi-treatment matching sample, we use the AUC index to judge the matching merit. The following formula calculates it.

\begin{equation}
    AUC=\frac{\sum  { pred }_{ {pos}}> { pred}_{ {neg }}}{{positiveNum * negativeNum }}
\end{equation}\par

It is noted that positive and negative cases may be unbalanced in this problem. In contrast, the calculation of AUC considers the classification ability of positive and negative cases and can reasonably evaluate the classifier despite the unbalanced sample. Therefore, AUC is not sensitive to whether the sample categories are balanced or not, and is suitable as a scoring criterion for this work.

\subsection{Ablation Study}
In this section, we test the effects of two techniques: optical enhancement and attention mechanism, on the model's performance before and after their introduction. Comparative experiments on multi-task loss will be shown later.
\subsubsection{Optical Enhancement Comparison}
We compared several optical enhancements mentioned in II. B. 2) and tested them in both datasets with the following results.\par
\begin{figure}[htbp]
\centering
\begin{minipage}[t]{0.48\textwidth}
\centering
\includegraphics[width=6cm]{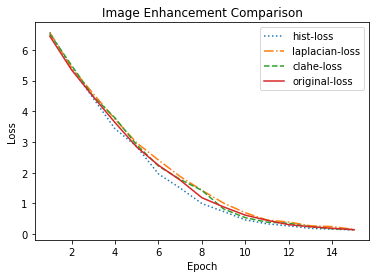}
\caption{Loss of different enhancement}
\label{figtp1}
\end{minipage}
\begin{minipage}[t]{0.48\textwidth}
\centering
\includegraphics[width=6cm]{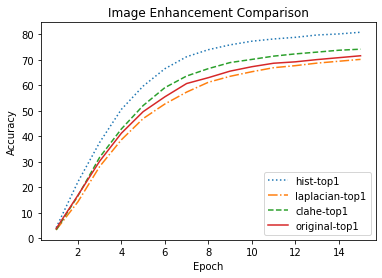}
\caption{Accuracy of different enhancement}
\label{figtp2}
\end{minipage}
\end{figure}

On the TJU-P dataset, the enhanced palm prints of the images by three different ways were compared with the results of the original images after 15 training sessions. And the batch size is 64, the learning rate is $5\times 10^{-5}$ and the optimiser is Adam. The results for loss and matching accuracy are shown in Figure \ref{figtp1} and Figure \ref{figtp2}.

From the experiments, it can be seen that after image enhancement by histogram equalization, loss converges the fastest and has the highest recognition rate improvement; CLAHE has the second highest effect; Laplace operator processing has the least significant effect, and the accuracy rate is even lower than the recognition rate of the unprocessed image.\par
On the PolyU M\_N palm vein dataset, only these two image enhancement methods were used for comparison as hist and CLAHE were shown to achieve better results through the above experiments. The parameters were chosen: the number of iterations was 15, the batch size was 8, and the learning rate and optimiser were the same as before. The comparison results are shown below in Figure \ref{figie1} and Figure \ref{figie2}.

\begin{figure}[htbp]
\centering
\begin{minipage}[t]{0.48\textwidth}
\centering
\includegraphics[width=6cm]{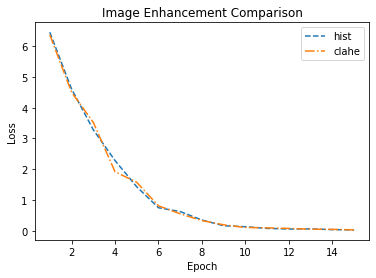}
\caption{loss}
\label{figie1}
\end{minipage}
\begin{minipage}[t]{0.48\textwidth}
\centering
\includegraphics[width=6cm]{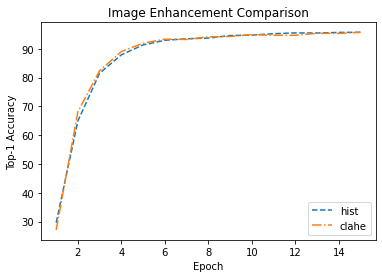}
\caption{accuracy}
\label{figie2}
\end{minipage}
\end{figure}
As can be seen from the figure: on the PolyU M\_N palm vein dataset, the enhancement effect of image enhancement using hist and clahe is almost the same, and hist is slightly better in terms of convergence speed and accuracy.
\subsubsection{Attention mechanisms}
\begin{figure}[htbp]
    \centering
\begin{center}
    \includegraphics[width = 0.7\linewidth]{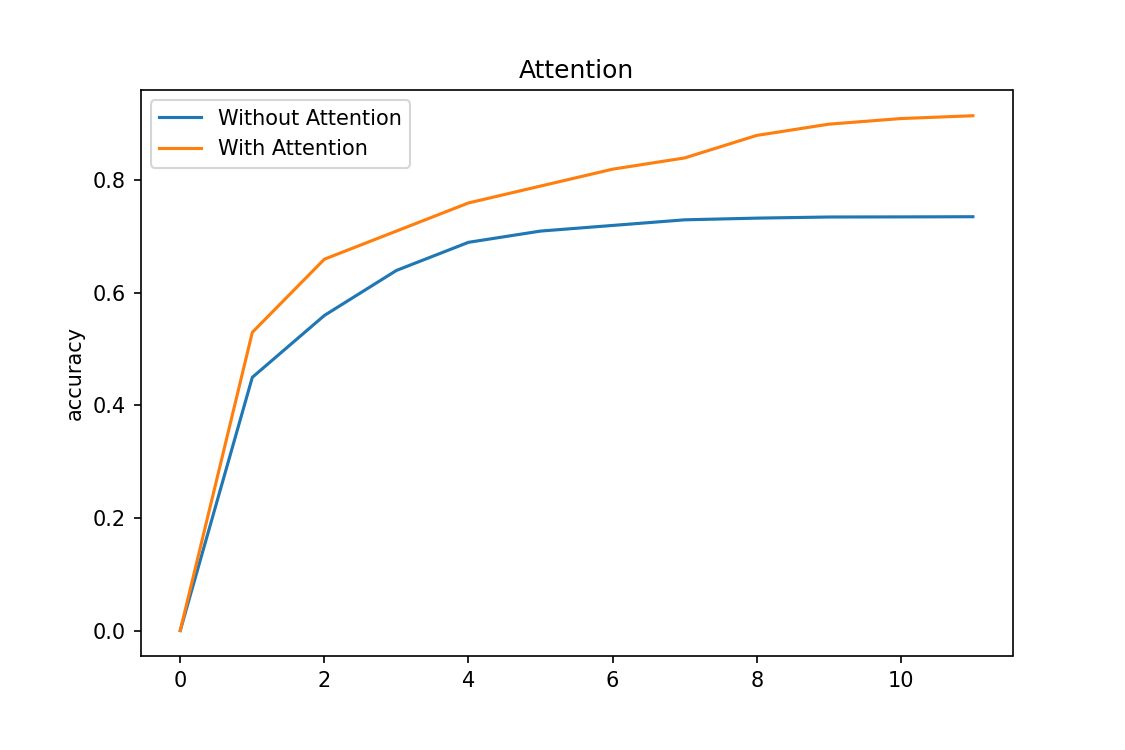}
\end{center}
    \caption{Effect of attention on accuracy}
    \label{acc}
\end{figure}

We then tested the effect of introducing attentional mechanisms on the experimental results. Using the Poly-U M\_N dataset as an example, we tested the matching accuracy with and without the inclusion of the attention mechanism. The results are presented in Figure \ref{acc}.\par
\par
The results show that the attention mechanism's introduction largely improves convergence speed. It also improves the accuracy of the model at convergence. \par
\begin{figure}[htbp]
    \centering
\begin{center}
    \includegraphics[width = 0.7\linewidth]{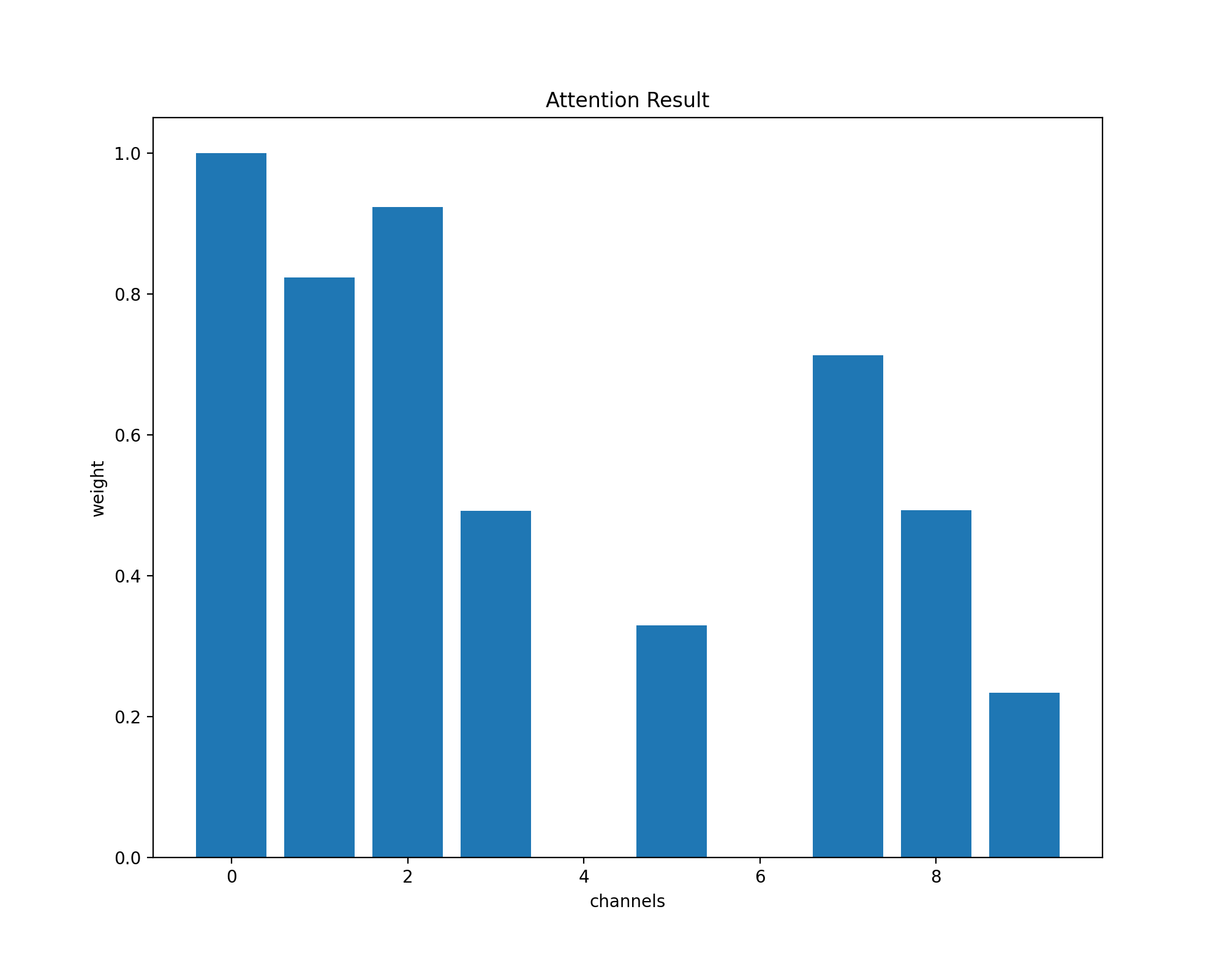}
\end{center}
    \caption{Effect of attention on accuracy}
    \label{attR}
\end{figure}

The attention mechanism can be used to weight different spatial regions or channels, as can be seen from the weighting of channels by the attention mechanism in the Figure \ref{attR}, where the weight of some channels is even set to 0. This also serves the purpose of feature selection. Since we found that the attention mechanism performed well in the model, we will use attention as the underlying structure in all subsequent experiments.\par

\par

\subsection{Results of Classification-matching Methods}
\subsubsection{Classification Results}
\begin{figure}[htbp]
\centering
\begin{minipage}[t]{0.48\textwidth}
\centering
\includegraphics[width=6cm]{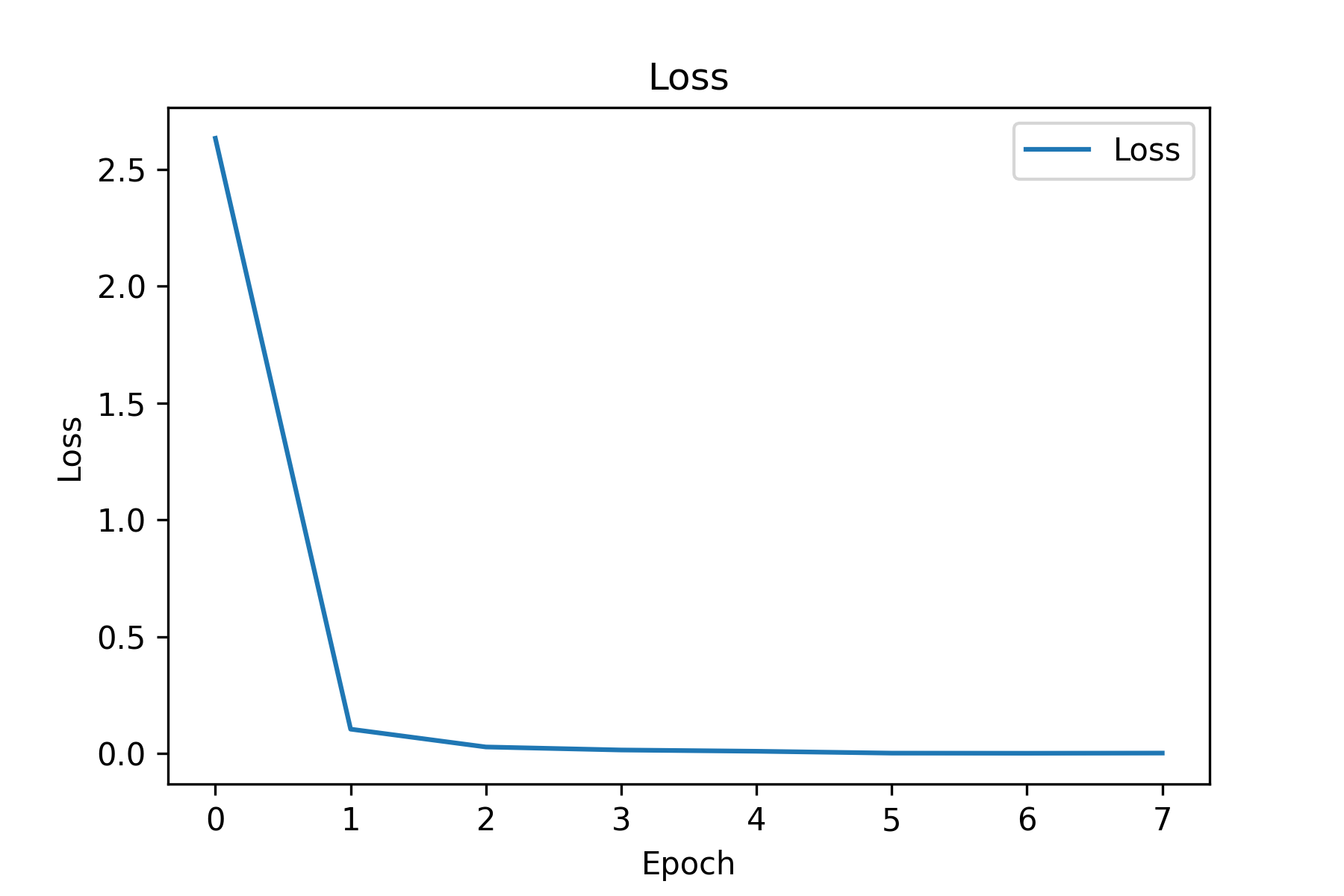}
\caption{Train loss}
\label{fig10}
\end{minipage}
\begin{minipage}[t]{0.48\textwidth}
\centering
\includegraphics[width=6cm]{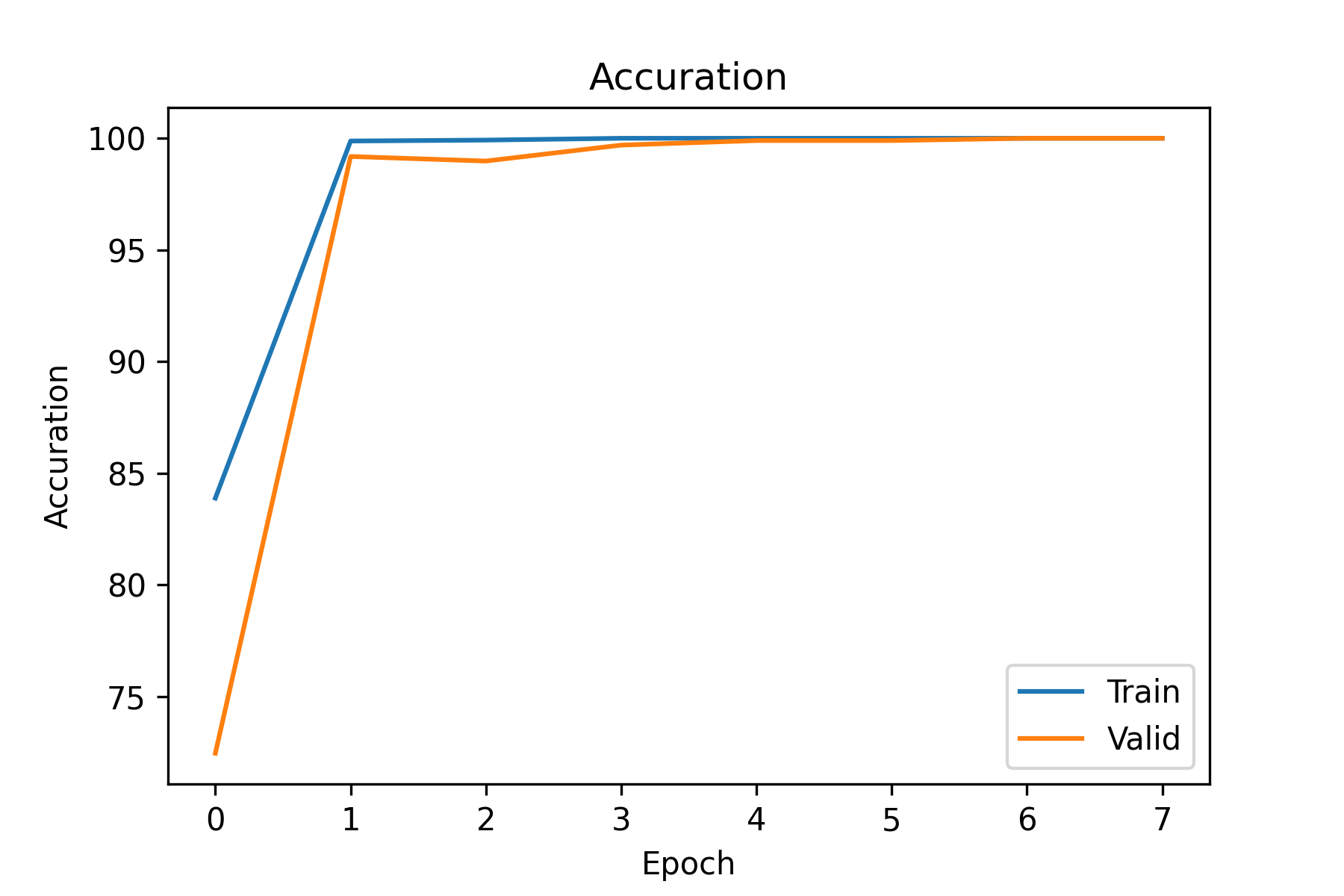}
\caption{Train accuracy}
\label{fig11}
\end{minipage}
\end{figure}

 Batch size = 32 is set, 20 rounds are trained, and a regularization strength of $\lambda = 0.001$ is used. The resulting classification accuracies and loss functions for the training set, validation set and test set are shown in Figure \ref{fig10} and Figure\ref{fig11}, and the training was stopped at the 7th round due to the fast convergence rate.

The training took a total of 173 seconds for 7 rounds, and it can be found from the figure that the model converged well. The accuracy is 100\% on the training set and 99.85\% on the validation set. We also tried different combinations of parameters, and the results are shown in the table 1.

\begin{table*}[htbp]
\centering
\begin{tabular}{cccccccc}
\hline
\textbf{Batch Size} & \textbf{Epoch} & \textbf{Regularization strength} & \textbf{Loss} & \textbf{Train Accuration} & \textbf{Test Accuration} & \textbf{Early stop} & \textbf{Training Time}\\ \hline
\textbf{32}         & \textbf{10}    & \textbf{0.001}                   & 0.000385      & 100\%                     & 99.85\%                  & 7   & 15m32s                \\
\textbf{64}         & \textbf{10}    & \textbf{0.001}                   & 0.000461      & 100\%                     & 98.73\%                  & No    & 12m14s               \\
\textbf{32}         & \textbf{5}     & \textbf{0.001}                   & 0.000612      & 99.63\%                   & 98.77\%                  & No        & 7m54s           \\
\textbf{64}         & \textbf{5}     & \textbf{0.001}                   & 0.001324      & 96\%                      & 95.87\%                  & No       & 5m59s            \\
\textbf{32}         & \textbf{10}    & \textbf{0.01}                    & 0.00836       & 95.42\%                   & 94.68\%                  & 8         & 15m17s           \\
\textbf{64}         & \textbf{10}    & \textbf{0.01}                    & 0.00972       & 95.11\%                   & 94.32\%                  & No        & 13m05s           \\
\textbf{32}         & \textbf{5}     & \textbf{0.01}                    & 0.0132        & 94.91\%                   & 89.23\%                  & No       & 8m16s            \\
\textbf{64}         & \textbf{5}     & \textbf{0.01}                    & 0.0187        & 91.61\%                   & 85.09\%                  & No       & 6m28s            \\ \hline
\end{tabular}
\label{tab1}
\caption{Experimental results of the original loss function}
\end{table*}

The results show that with the regularization strength of $\lambda = 0.001$, the final loss and accuracy are not sensitive to the number of batches and training sessions. However, for $\lambda = 0.01$ the model appears to be underfitted.\par
At the same time, we found that too large batch size will greatly affect the training effect. Although a large batch size helps to speed up training, with the support of a powerful GPU, the time consumption here is negligible.

\subsubsection{Matching Results}
\begin{figure}[H]
\centering
\begin{minipage}[t]{0.48\textwidth}
\centering
\includegraphics[width=6cm]{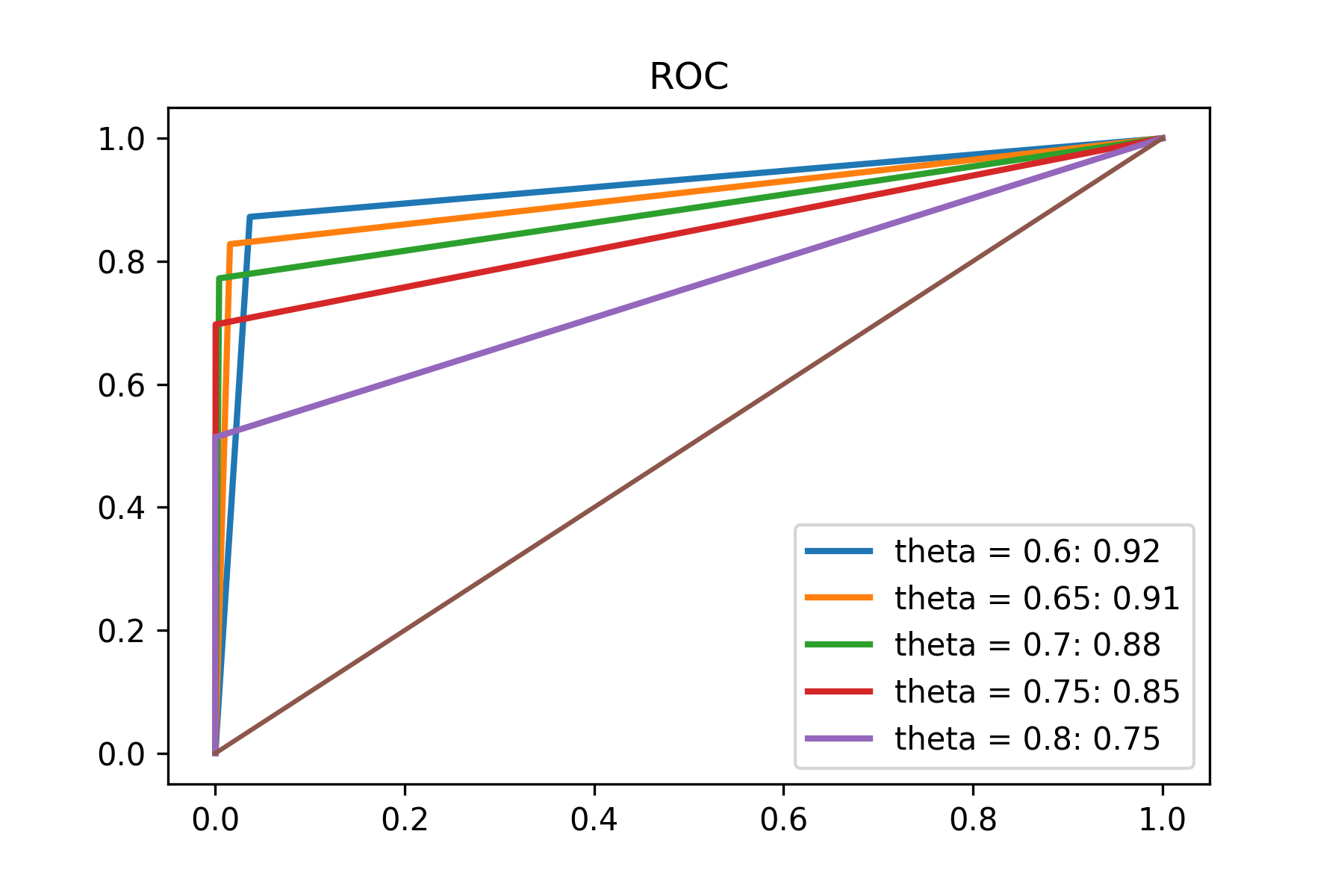}
\caption{Comparison of AUC at different thresholds}
\label{pic1}
\end{minipage}
\end{figure}

\begin{figure}[H]
\centering
\begin{minipage}[t]{0.48\textwidth}
\centering
\includegraphics[width=6cm]{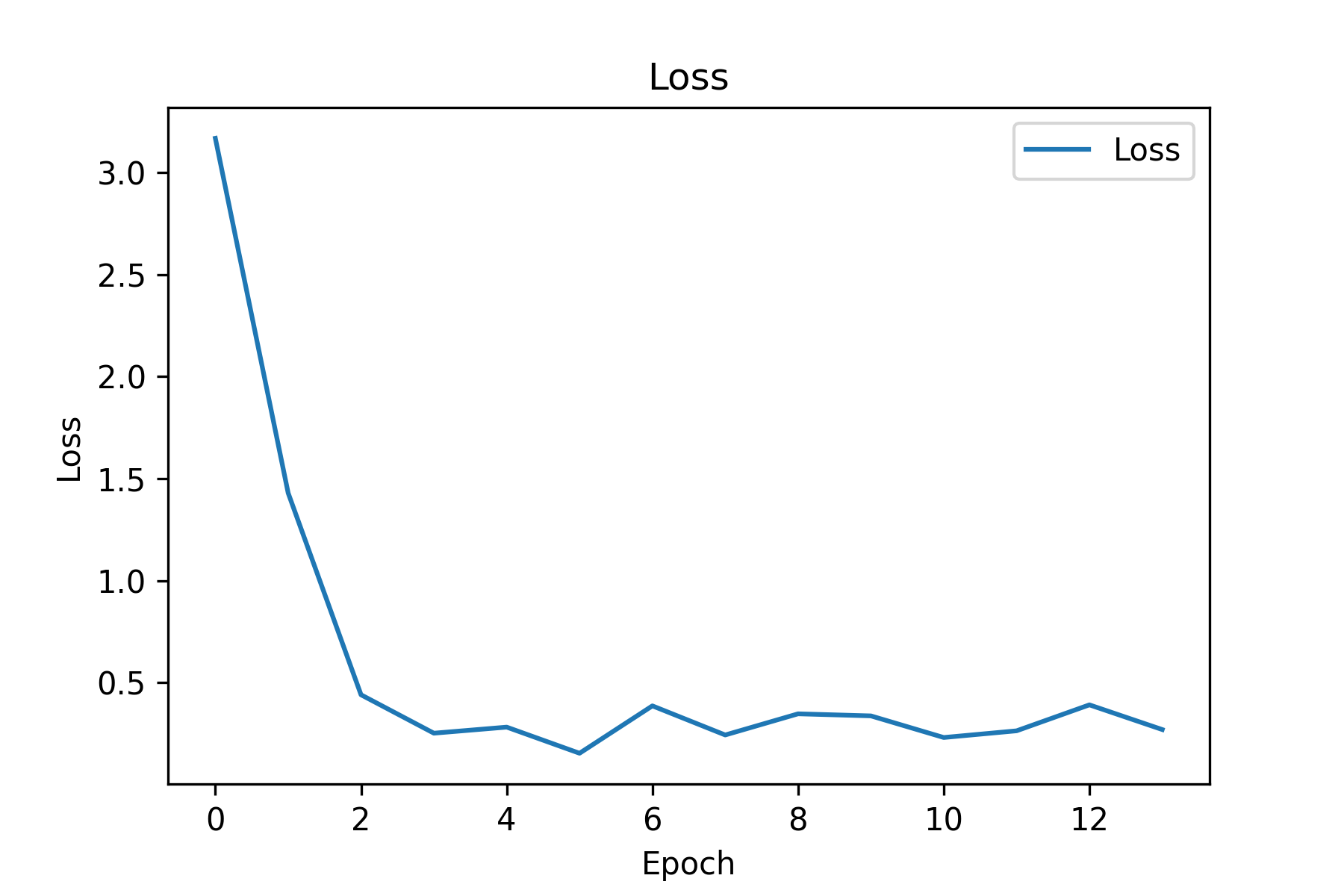}
\caption{multi-tasks loss function loss}
\label{pic2}
\end{minipage}
\begin{minipage}[t]{0.48\textwidth}
\centering
\includegraphics[width=6cm]{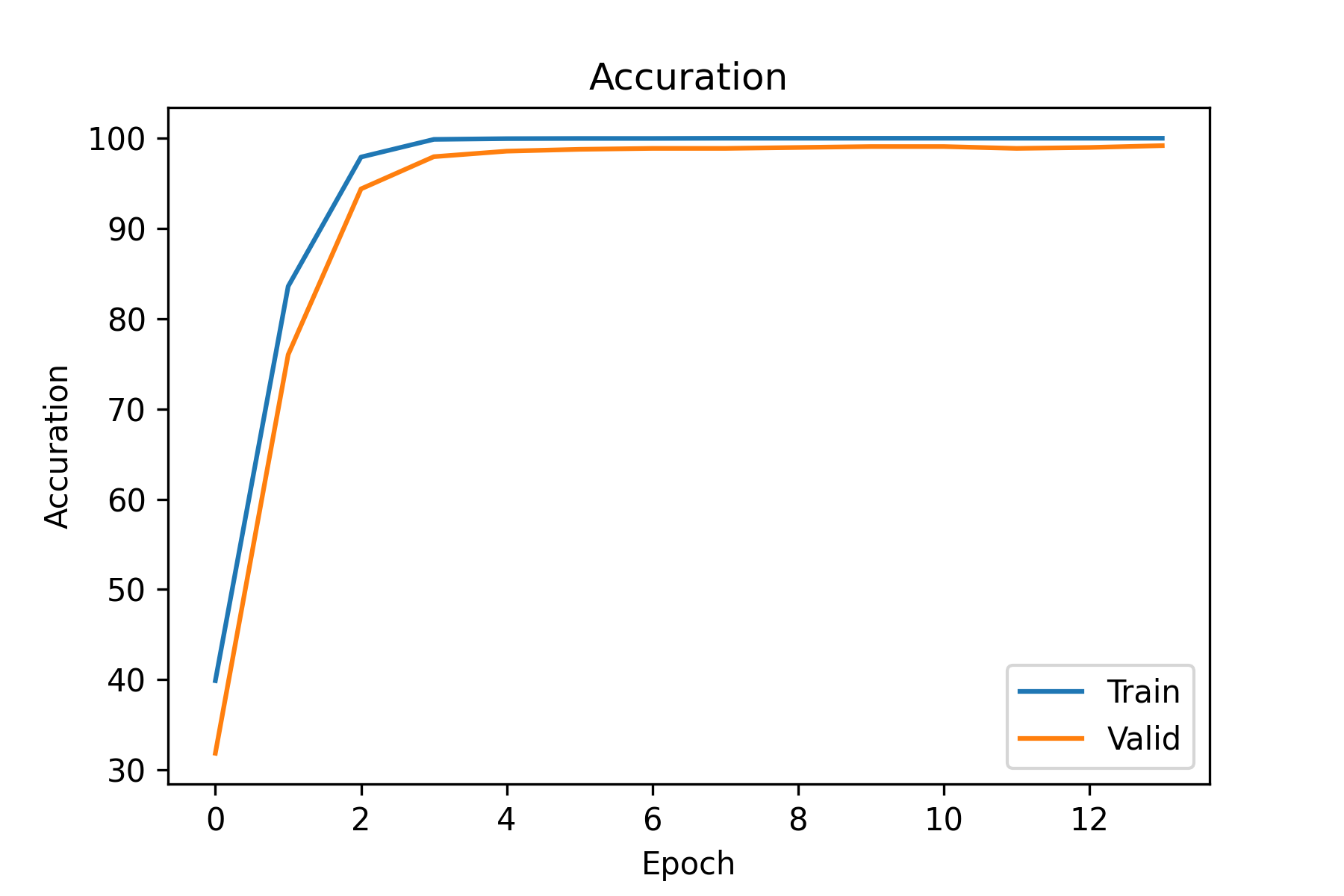}
\caption{multi-tasks loss function accuracy}
\end{minipage}
\end{figure}

Next, the matching effect is tested on the dataset $TE_m$. Also, to get a better matching effect, we test thresholds of 0.6, 0.65, 0.7, 0.75, and 0.8, calculate their AUCs and compare them. Figure \ref{pic1} shows the comparison graph of AUC, and it can be found that the best result is obtained when the threshold is 0.6. Figure \ref{pic2} shows the prediction when the threshold is 0.6, where yellow is the correct prediction and blue is the wrong prediction.

The overall correct rate was 95.08\%. Among them, 279 pairs were correctly accepted, 121 pairs were incorrectly rejected, 56 were incorrectly accepted, and 3144 were correctly rejected. Although the overall accuracy rate is very high, the acceptance rate for positive samples is very low, even less than 50\%. Considering that the classification accuracy has reached 100\%, we need to modify the loss function in the training to further improve the matching accuracy.\par
Meanwhile, we used the (\ref{equ8}) equation to calculate the goodness-of-fit under the matching task, i.e., the difference between the average similarity from the same sample and the average similarity from different samples were calculated separately. The experimental results show that the average similarity of samples from the same palm is 0.7700, and the average similarity of samples from different palms is 0.2531, with a difference of 0.5169, which is the $MG$. We find that the correct acceptance rate of this method is low and far from the standard for practical use, so we will try our improved objective function next. \par

\subsection{Result based on multi-tasks loss function}

Set Batch size = 128, keep the rest parameters the same as the optimal parameters above, change the value of $\theta$ for several experiments, and take one of the training accuracy and loss variation graphs to show as follows.

We found that the model converged slower due to the improvement of the loss function. The model stopped training at 12 rounds and took 1247 seconds. The loss function's training time is higher than the normal classification task. We tried different combinations of parameters and the obtained experimental results are shown in the table 2.

\begin{table*}[htbp]
\centering
\begin{tabular}{ccccccccc}
\hline
\textbf{Theta} & \textbf{Alpha} & \textbf{Correctly accepted} & \textbf{Wrong accepted} & \textbf{Wrong rejected} & \textbf{Correctly rejected} & \textbf{Correct Rate} & \textbf{AUC} & \textbf{MG} \\ \hline
\textbf{0.5}   & \textbf{0.7}   & 310                         & 132                     & 36                      & 3122                        & 95.33\%               & 0.9442 & 0.7942       \\
\textbf{0.3}   & \textbf{0.7}   & 345                         & 79                      & 15                      & 3161                        & 97.39\%               & 0.9669  & 0.8293     \\
\textbf{0.1}   & \textbf{0.7}   & 324                         & 79                      & 36                      & 3161                        & 96.81\%                & 0.9378  & 0.8121     \\ \hline
\end{tabular}
\caption{Experimental results of multi-tasks loss function}
\end{table*}

Where $\alpha$ are used to select the optimal value by the above method and change the value of $\theta$, the results show that the highest matching accuracy is achieved when $\theta = 0.3$, when the loss function is as follows.

\begin{equation}
   loss = 0.3* Crossentropy(X, y)  +0.7*(1-MG) + \lambda\left \| w \right \|_2
    \label{equ14}
\end{equation}\par
The following Figure \ref{or} and Figure \ref{multi} shows the experimental results of the original loss function and the multi task loss function on the matching task. The yellow data points represent correctness and the blue data points represent errors.

\begin{figure}[H]
\centering
\begin{minipage}[t]{0.48\textwidth}
\centering
\includegraphics[width=6cm]{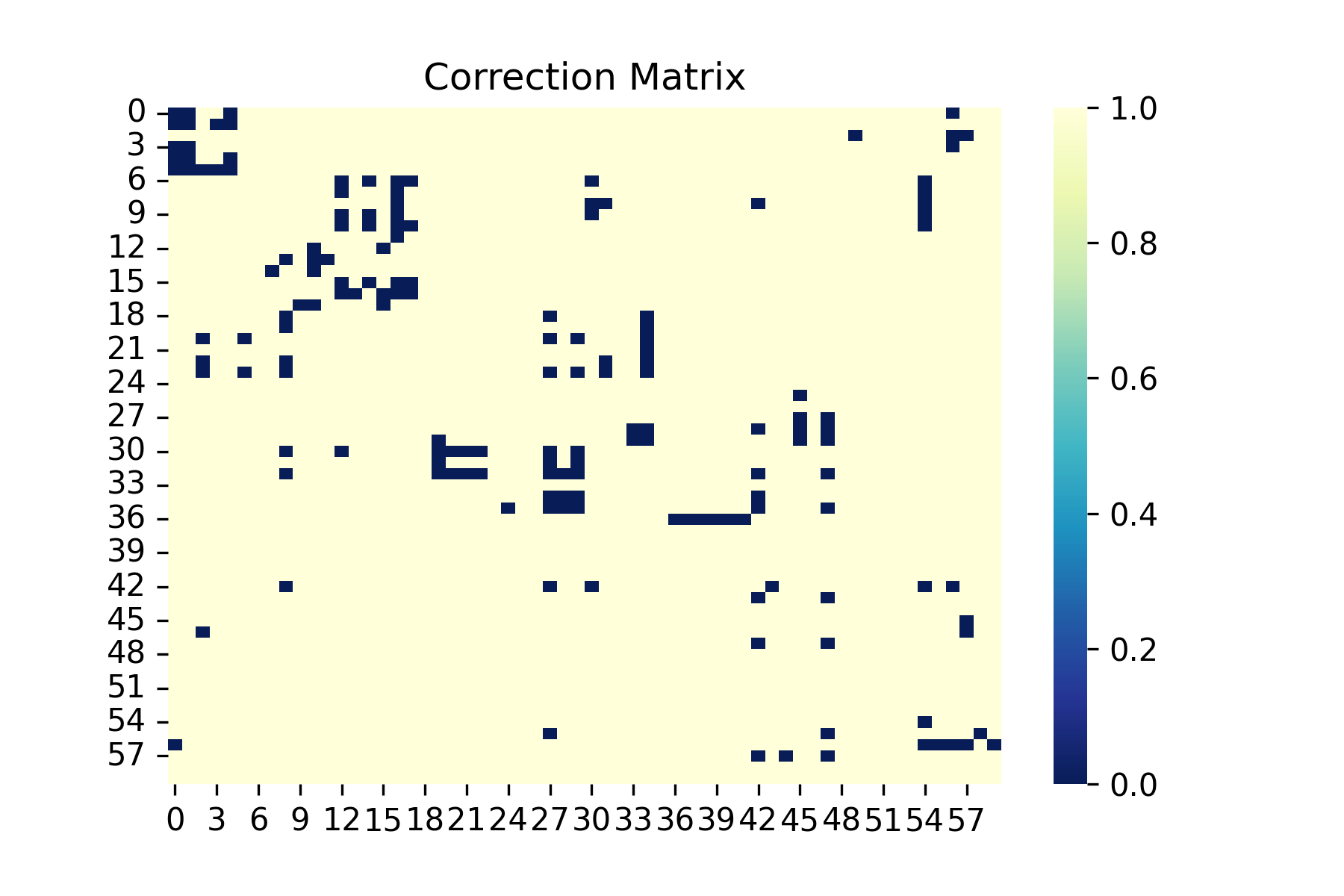}
\caption{Original loss function prediction results}
\label{or}
\end{minipage}
\begin{minipage}[t]{0.48\textwidth}
\centering
\includegraphics[width=6cm]{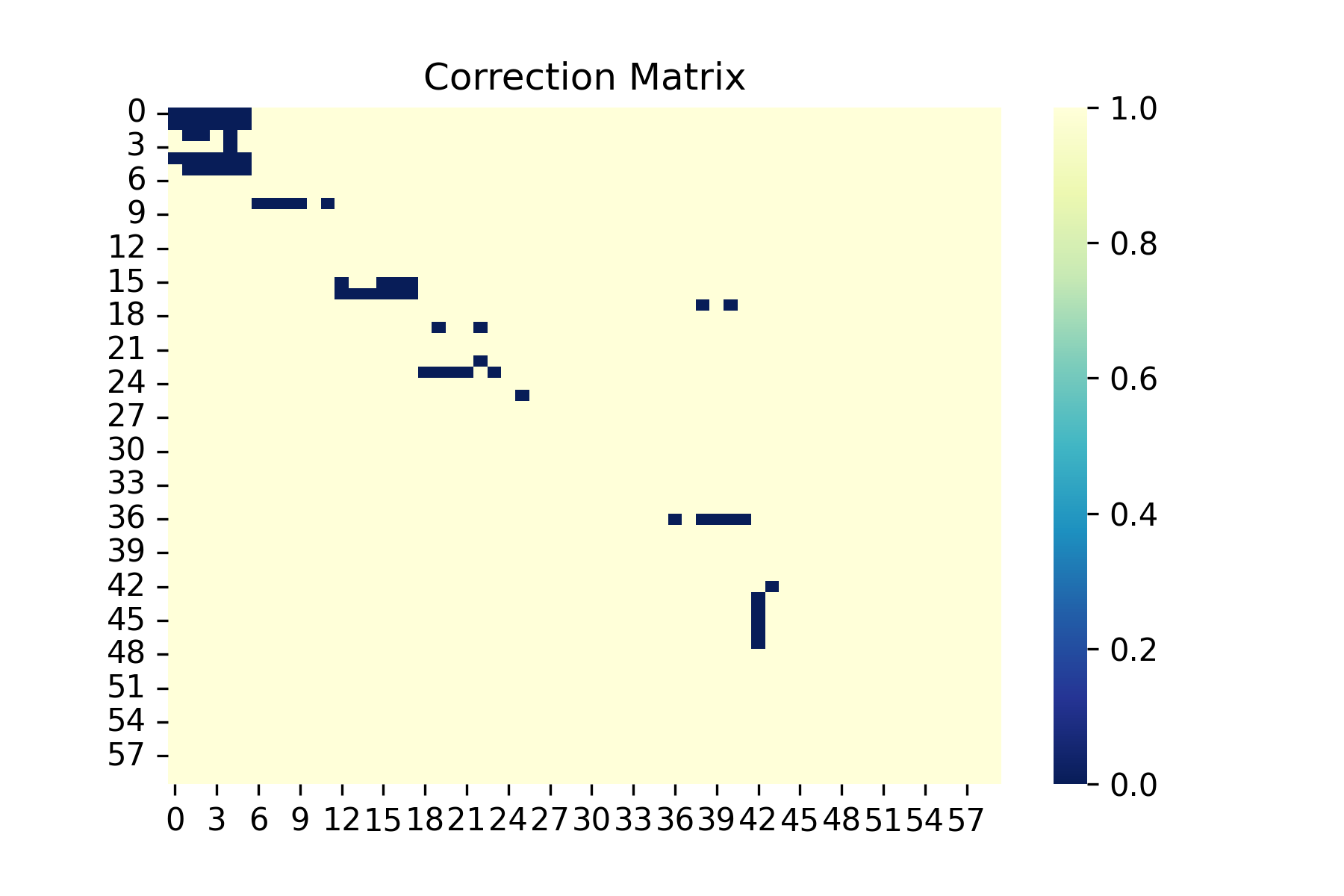}
\caption{multi-tasks loss function prediction results}
\label{multi}
\end{minipage}
\end{figure}\par

It can be seen that the multi-task loss function proposed in this paper has a higher prediction accuracy. We also test the goodness of matching, the difference between the average similarity of samples from different palms and the same palms. It is 0.8293 which is far higher than the result in section A.

\subsection{The case where the training and test sets are from different datasets}
The palm vein images used in the above model training and matching are from the same dataset. In other words, the locations and parameters of the photos taken are approximately the same. Testing these photos alone does not measure the true robustness and transferability of the model. Therefore, we next use photos taken in different ways for matching tests.
To improve the model's accuracy, we use palm vein images taken in two completely different environments for training, and use the dataset from the third case for matching tests. The appearance of the training image and the test image are shown below.

\begin{figure}[htbp]
    \centering
\begin{center}
    \includegraphics[width = \linewidth]{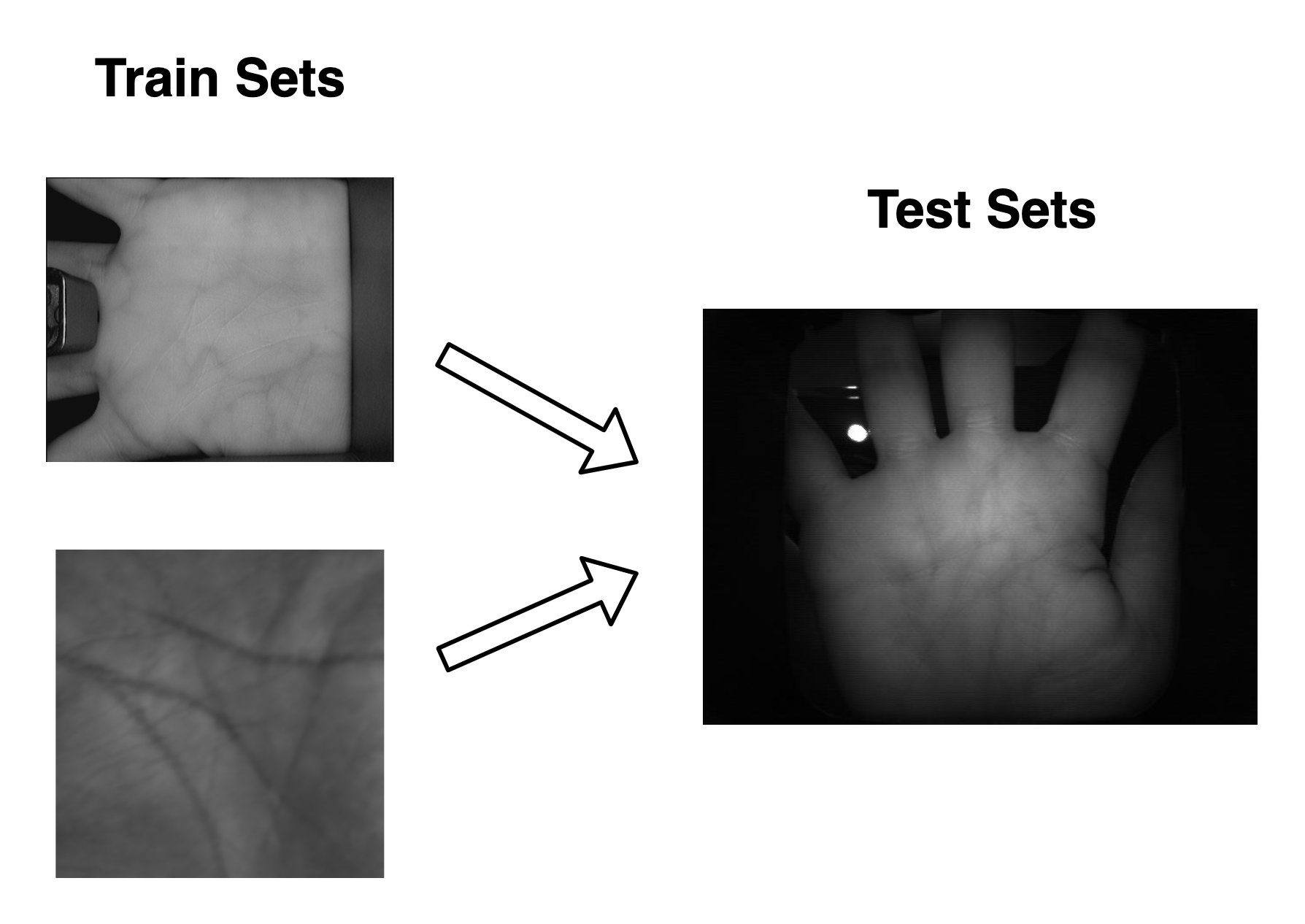}
\end{center}
    \caption{Different training and test sets}
\end{figure}\par

The palm veins are more informative in the top left image, while the palm surface is more informative in the bottom left image. And in this case, the determination of the matching threshold is key. Since the test and training sets are from very different datasets, the positive and negative examples cannot be simply partitioned by a number like 0.6 or 0.7. Therefore, we must try to determine the matching threshold $\alpha$. From the above 
discussion, we can find that based on the Multi-tasks loss function, we believe that the final set of obtained feature vectors should satisfy the $MG$(\ref{equ8}) minimum. In this case, we believe that the similarity of positive and negative examples should each be well separated, and although they will partially cross, we can always find a better threshold based on such an assumption. Therefore, we first perform K-mean\cite{ref45} clustering on the similarity of all samples and get two clustering centers $p$ and $n$, and we take $\alpha = \frac{p+n}{2}$ as the threshold for matching.\par

In the case of the experiments in this paper, we finally choose $\alpha = 0.992$. Predictions were performed for a test set with 46 pairs of positive cases and 134 pairs of negative cases. Out of 180 pairs of predictions, only 2 pairs of palms that should have belonged to the same person were incorrectly rejected. The overall accuracy rate was 98.89\%. The following figure shows the confusion matrix of the predicted results.

\begin{figure}[htbp]
    \centering
\begin{center}
    \includegraphics[width = \linewidth]{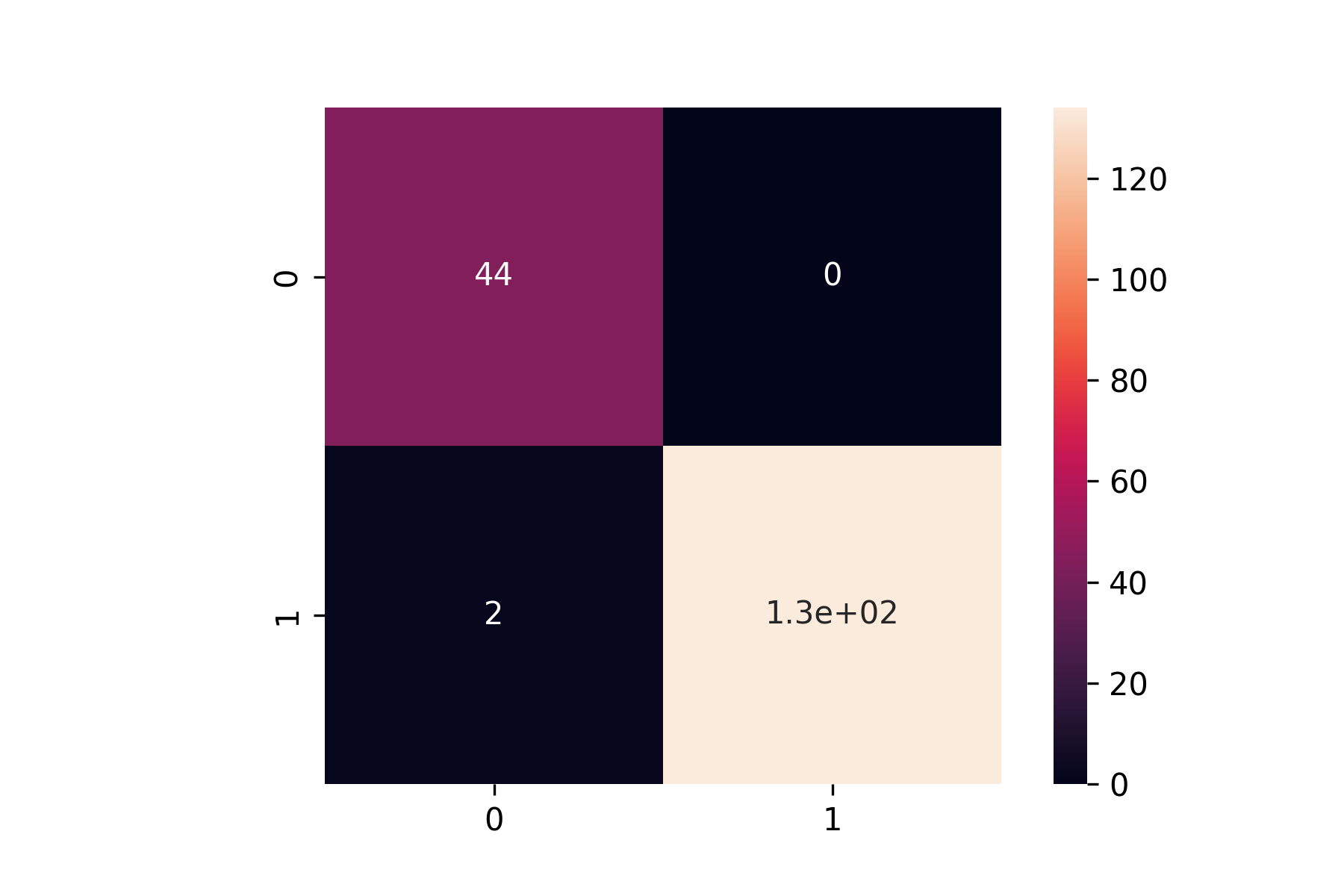}
\end{center}
    \caption{Confusion matrix of test results}
\end{figure}\par
We find that the model still shows very high accuracy on different datasets, so we consider the training idea of the palm vein matching model we have provided is relocatable. At the same time, we compared the accuracy of the validation set under different methods and the results obtained are shown in the Table 3. The results show that the VGG network with attention mechanism and multi-task loss improves 11.36\% in accuracy over the original VGG network, and 5.11\% over the attention mechanism alone. \par

\begin{table*}[htbp]
\centering
\begin{tabular}{ccccccccc}
\hline
 \textbf{Model} & \textbf{Correctly accepted} & \textbf{Wrong accepted} & \textbf{Wrong rejected} & \textbf{Correctly rejected} & \textbf{Correct Rate} & \textbf{AUC} & \textbf{MG}\\ \hline
 \textbf{VGG}   & 115                         & 7                     & 15                      & 39                        & 87.50\%               & 0.8662    & 0.7301   \\
 \textbf{Attention-VGG}   & 124                         & 5                      & 6                      & 124                        & 93.75\%               & 0.9226    & 0.8271   \\
 \textbf{multi-task-Attention-VGG}   & 130                         & 2                      & 0                      & 44                        & 98.86\%                & 0.9783       & 0.8724\\ \hline
\end{tabular}
\label{tb3}
\caption{Accuracy comparison on the validation set}
\end{table*}

\subsection{Simulation of reality: noise, displacement and low resolution}

\begin{figure}[H]
    \centering
\begin{center}
    \includegraphics[width = \linewidth]{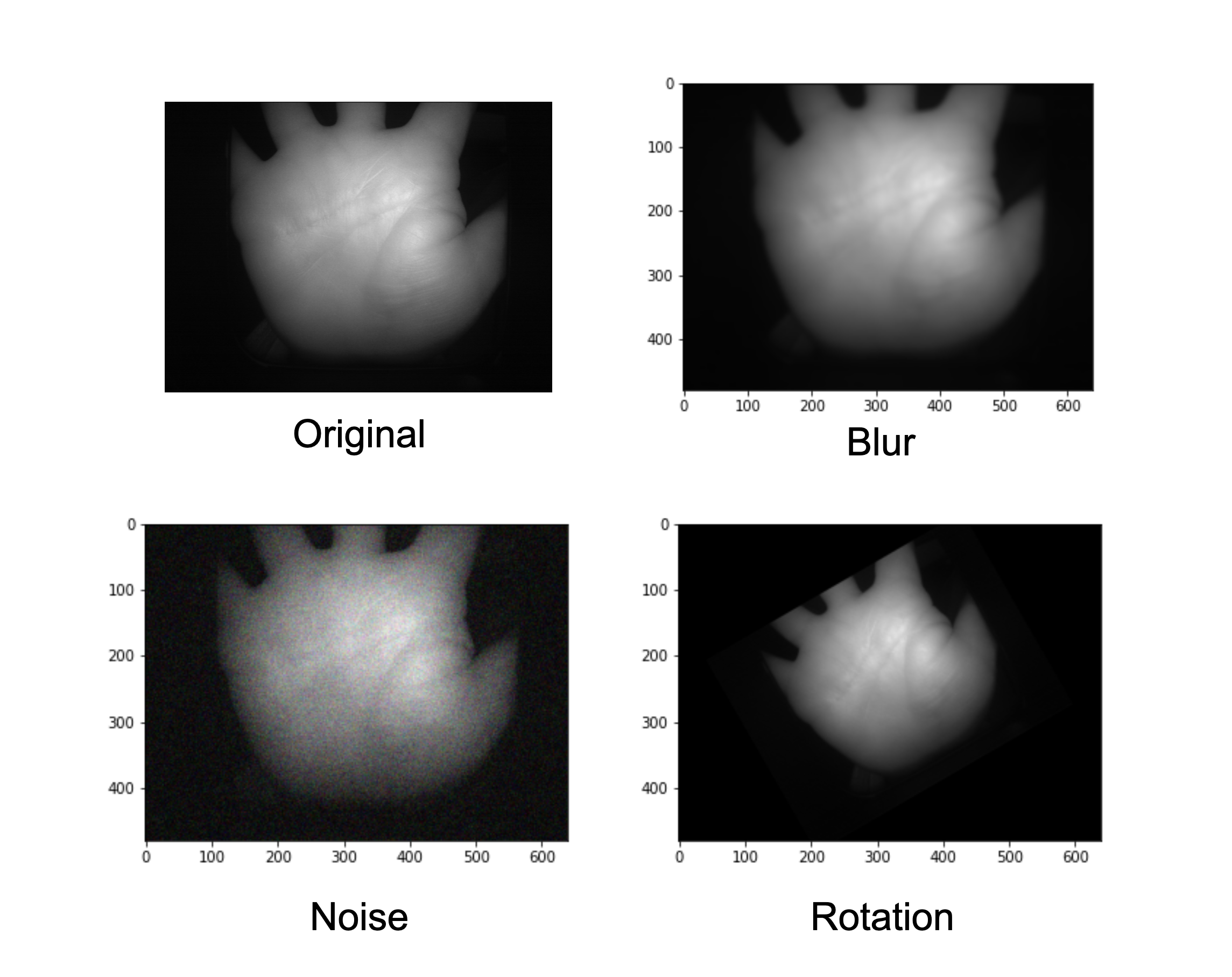}
\end{center}
    \caption{Simulation of realistic image processing}
    \label{real}
\end{figure}

The palm print dataset in the above experiments was collected in a very regular pattern and they used a limiter to fix the position of the palm. However, in real-life application, the position of the hand, the environment, and the resolution of the acquisition device may all change. Therefore, we rotate, blur and randomly noise the palmprint images of the validation set. And experiments are conducted in the same way as above to explore the robustness of the model. Figure \ref{real} shows a sample of images with the four processing methods

\par

We did this for each image in the validation set and compared the AUC of the different treatments. The results are shown in Table 4.

\begin{table}[H]
\centering
\begin{tabular}{cccc}
\hline
\textbf{Treatment}         & \textbf{Correct Rate} & \textbf{AUC} & \textbf{MG}     \\ \hline
\textbf{Original} & 98.86\%               & 97.83\%      & 0.8724 \\
\textbf{Blur}     & 57.16\%               & 53.44\%      & 0.4139 \\
\textbf{Noise}    & 93.72\%               & 92.73\%      & 0.8322 \\
\textbf{Rotation} & 92.14\%               & 91.09\%      & 0.8147 \\ \hline
\end{tabular}
\caption{Results of experiments in simulating realistic scenarios}
\end{table}
We can find that matching accuracy decreases in all three unique treatments. The Noise and Rotate drop is insignificant and still has an accuracy rate of over 90\%. On the other hand, the model has almost no recognition ability for the blur. Our analysis is that this is due to blurring resulting in the complete loss of otherwise obscure palm print features, which makes samples from the same individual look the same as samples from different individuals. This is illustrated by MG, which is only 0.4139. This indicates that the similarity of the two types of samples has a considerable overlap, which means that a threshold cannot simply split it.\par
The above experimental results show that this method is robust to both common noise and rotation. The loss of features due to blurring means that there is a requirement for the device's resolution in this recognition mode.
\subsection{Time consumption}

\begin{figure}[htbp]
    \centering
\begin{center}
    \includegraphics[scale = 0.5]{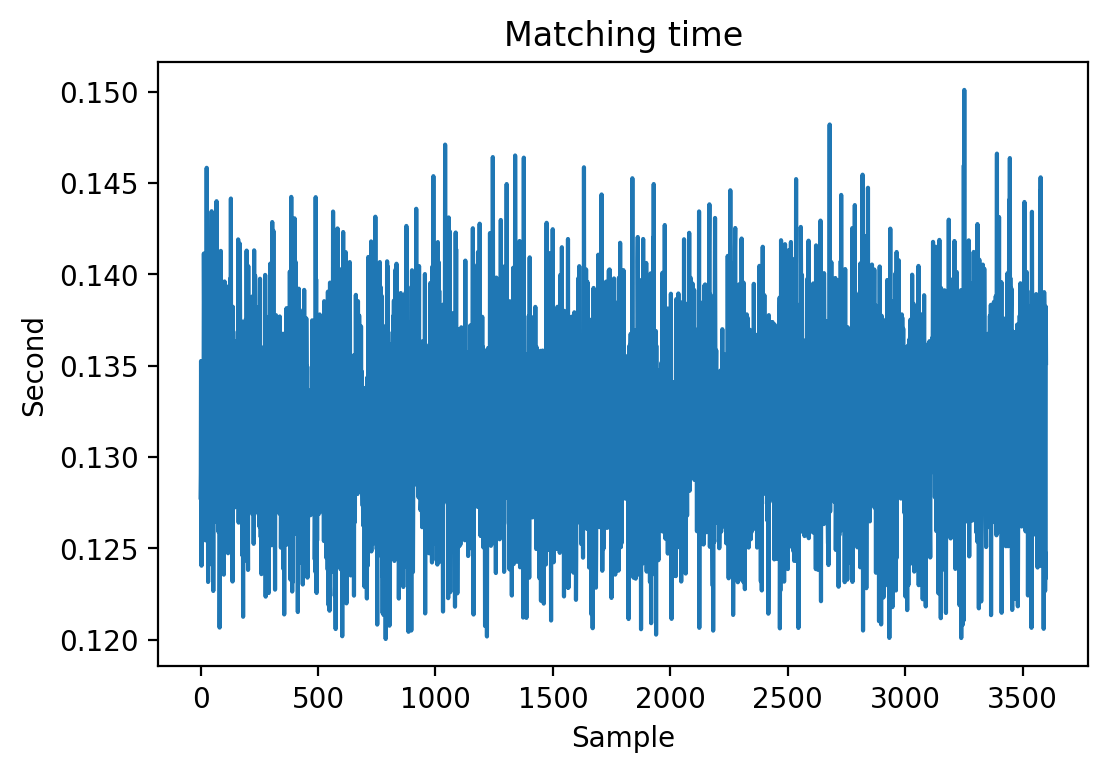}
\end{center}
    \caption{Prediction time consumption}
    \label{time_use}
\end{figure}

We also test the time complexity of the matching process. Considering that most of the application scenarios are not supported by GPU arithmetic, in order to simulate real usage scenarios, the test will be based on the Apple M1 Pro platform using CPU for inference. In the test on 3600 pairs of samples, the average test time per pair of samples was 0.1316 seconds, and the time consumption curve for each pair of samples is shown in Figure \ref{time_use}.

\par

The results show that the algorithm is short time consuming and stable, which has the possibility to be used in real life.

\section{Conclusion, Discussion and Future Work}

\subsection{Innovation}

In this paper, we combine optical feature extraction with depth feature extraction from the traditional method of optical recognition of palm veins. Through experiments, it is found that this approach can better mine image texture features and provide features that can be multi-dimensional for matching analysis. At the same time, the use of VGG-16 transfer learning improves the learning efficiency and makes the model training overhead much lower; the practice of freezing part of the pre-training parameters makes use of the original image recognition ability of the network on the one hand and can be better adapted to different tasks on the other. At the same time, we introduce the Attention mechanism, which enables the model to distinguish between primary and secondary features after feature extraction, improving the model robustness and further increasing the model accuracy. Finally, we use the trained model for the palm vein matching task and get an accuracy rate of more than 98\%. \par

At the same time, we propose an index to evaluate the training results' average Similarity, which can represent the distance of similarity between different kinds of palmprint images, to describe the accuracy and difficulty of distinguishing them. In the future study, we can use this metric to evaluate and compare traditional matching solutions, such as using PCA to extract features by dimensionality reduction and using the SVM support vector machine to match \cite{ref41} \par

Finally, to satisfy the migration of the model to other datasets. We determine the adaptive matching threshold based on a clustering algorithm, and the experimental results prove that the method is simple and feasible with high accuracy.\par

\begin{figure}[H]
    \centering
\begin{center}
    \includegraphics[width = \linewidth]{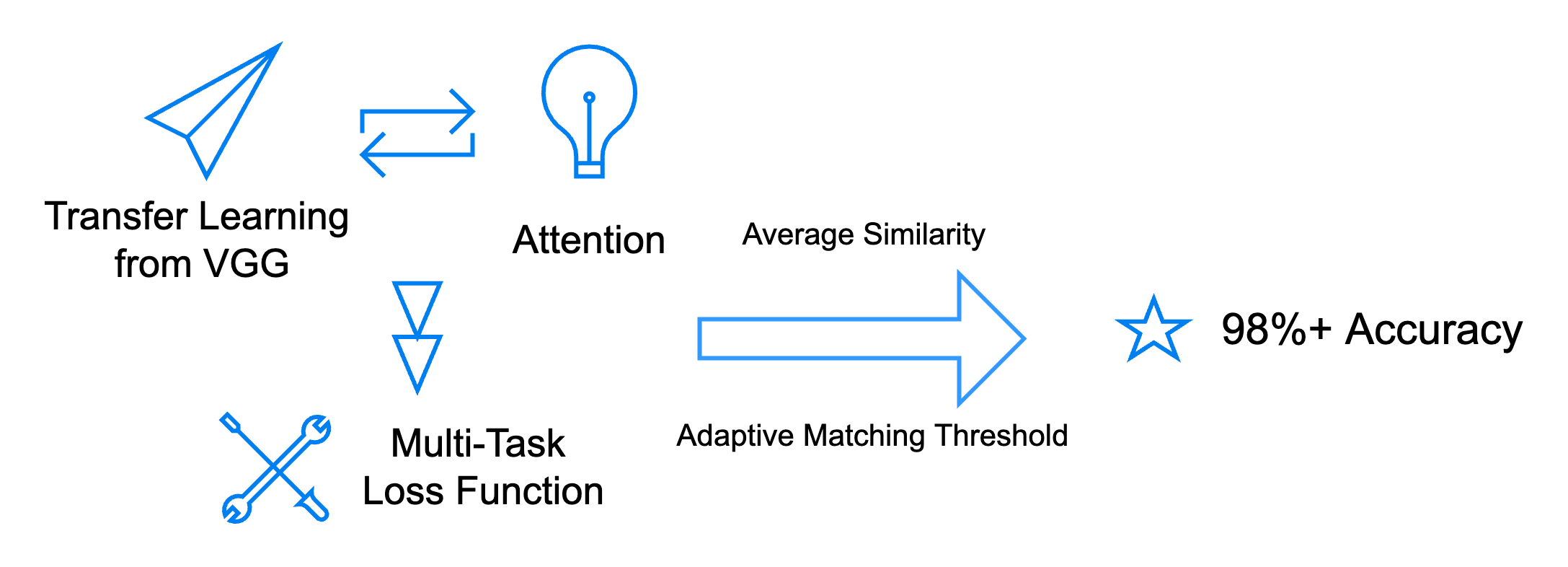}
\end{center}
    \caption{Innovations}
\end{figure}\par

\subsection{Shortcomings and future work}

However, in this paper, no special treatment is done for the dataset. A random method is still used for the division of batches. In future work, we can reconstruct the dataset so that the number of data from the same sample and different samples in each batch is approximately the same. This would be more representative and reliable for the $MG$ calculation, thus further improving the model performance. Meanwhile, this paper's VGG network used for migration learning is not the best image recognition network available. There are well-performing networks such as Resnet. The latest results are given by NFNet\cite{ref44} proposed by DeepMind.\par

A similar improvement to this idea was proposed by Google at CVPR 2015 \cite{ref37}, which reconstructs the face classification dataset and proposes Triplets Loss for training. This network is called FaceNet, and its ultimate goal is to embed face images into a 128-dimensional Euclidean space and get a better matching result by minimizing the Euclidean distance between individuals and maximizing the Euclidean distance between different individuals. \par

Furthermore, due to the lack of infrared devices, this paper has not yet conducted demo production and application tests on the matching of the model, i.e., its performance in continuous and real-time recognition has not been tested. These aspects need to be further explored and studied.

\EOD

\end{document}